\documentclass{article}

\usepackage{microtype}
\usepackage{graphicx}
\usepackage{subfigure}
\usepackage{booktabs} 
\usepackage{float}
\usepackage{multirow}
\usepackage{threeparttable}
\usepackage{minitoc}

\usepackage{hyperref}

\usepackage[accepted]{icml2024}

\usepackage{amsmath}
\usepackage{amssymb}
\usepackage{mathtools}
\usepackage{amsthm}

\usepackage[capitalize,noabbrev]{cleveref}

\theoremstyle{plain}

\theoremstyle{definition}

\theoremstyle{remark}

\usepackage[textsize=tiny]{todonotes}

\icmltitlerunning{Enhancing Multivariate Time Series Forecasting by Constructing Auxiliary Time Series}

\begin{document}

\twocolumn[
\icmltitle{CATS: Enhancing Multivariate Time Series Forecasting by Constructing Auxiliary Time Series as Exogenous Variables}



\icmlsetsymbol{equal}{*}

\begin{icmlauthorlist}
\icmlauthor{Jiecheng Lu}{gt}
\icmlauthor{Xu Han}{aws}
\icmlauthor{Yan Sun}{gt}
\icmlauthor{Shihao Yang}{gt}
\end{icmlauthorlist}

\icmlaffiliation{gt}{Georgia Institute of Technology}
\icmlaffiliation{aws}{Amazon Web Services}

\icmlcorrespondingauthor{Jiecheng Lu}{jlu414@gatech.edu}
\icmlcorrespondingauthor{Shihao Yang}{shihao.yang@isye.gatech.edu}

\icmlkeywords{Machine Learning, ICML}

\vskip 0.3in
]



\printAffiliationsAndNotice{}  

\begin{abstract}
For Multivariate Time Series Forecasting (MTSF), recent deep learning applications show that univariate models frequently outperform multivariate ones. To address the deficiency in multivariate models, we introduce a method to \underline{C}onstruct \underline{A}uxiliary \underline{T}ime \underline{S}eries (CATS) that functions like a 2D temporal-contextual attention mechanism, which generates Auxiliary Time Series (ATS) from Original Time Series (OTS) to effectively represent and incorporate inter-series relationships for forecasting. Key principles of ATS—continuity, sparsity, and variability—are identified and implemented through different modules. Even with a basic 2-layer MLP as the core predictor, CATS achieves state-of-the-art, significantly reducing complexity and parameters compared to previous multivariate models, marking it as an efficient and transferable MTSF solution. The code implementation is available at this  \href{https://github.com/LJC-FVNR/CATS}{\underline{link}}
\end{abstract}

\section{Introduction}\label{introduction}
Time series forecasting plays a pivotal role in domains like finance, health, and traffic, focusing on predicting future data based on historical records \citep{kaastra1996designing,nihan1980use, sun2023manifold, analytics1020014}. The complexity of multivariate time series forecasting (MTSF) increases due to the need for understanding dependencies among multiple series. Effectively modeling these relationships is key to overcoming challenges such as overfitting, thus enhancing the reliability of predictions.

Recent research in MTSF with deep learning methods reveals a counter-intuitive trend: univariate models, which ignore inter-series relationships, often outperform their multivariate counterparts \citep{zeng2022transformers, nie2023time, lu2023arm}. This phenomenon suggests two potential issues: 1) The real-world benchmark datasets selected for study may inherently possess weak inter-series relationships; 2) Previous models may lack the proper structure needed to effectively capture inter-series relationships. This inadequacy often leads to overfitting or fitting of wrong patterns. Moreover, in situations where intra-series (temporal) relationships dominate, attempts to incorporate inter-series information may inadvertently disrupt the learning of temporal dynamics, resulting in multivariate underperformance. This highlights the need for models that can balance and harness both types of relationships in MTSF.

In this paper, we introduce an approach to effectively capture the two types of relationships in MTSF: \underline{C}onstructing \underline{A}uxiliary \underline{T}ime \underline{S}eries (CATS). This approach empowers a time series predictor to handle diverse strengths of multivariate relationships without altering its model architecture, even when the predictor is a structurally simple univariate model. CATS operates by generating Auxiliary Time Series (ATS) based on the Original Time Series (OTS), serving as exogenous variables to represent the inter-series relationships. By predicting these exogenous time series, CATS finds and incorporates multivariate information to the prediction of OTS, thereby boosting overall performance.

We define ATS as the intermediary time series that are derived from the OTS input, reside in the same temporal domain as the original series. ATS can refine the univariate forecasting results by incorporating inter-series information. There are three key principles of ATS: continuity, sparsity, and variability. These principles are crucial when the inter-series relationships are weak, complex, or challenging to learn, and allow the model to adaptively adjust itself to suit datasets with varied characteristics. ATS are constructed with these three principles using specific neural network structures, gating operations, and loss terms. This construction allows the ATS to effectively extract and represent multivariate relationships.

The key contributions of this paper can be summarized as: 

1) We introduce the CATS approach, which constructs ATS based on the multivariate OTS and empowers time series predictors to efficiently capture inter-series relationships through a 2D attention-like mechanism.

2) We identify three key principles—continuity, sparsity, and variability—that enhance the performance and flexibility of ATS, enabling adaptive and accurate representation of inter-series dependencies and facilitating correct pattern fitting across various datasets.

3) We propose various types of ATS constructors, offering adaptability to general MTSF datasets, demonstrating the versatility of CATS in handling diverse MTSF challenges.

\section{Related Works}

Time series forecasting models have been extensively studied in traditional statistical literature \citep{box1974some,holt2004forecasting,VAROriginal}. The advent of deep learning methods has provided time series forecasting with enhanced fitting capabilities, especially those based on RNN and CNN \citep{hochreiter1997long, wen2017multi, rangapuram2018deep, salinas2020deepar, qin2017dual, borovykh2017conditional, vanwavenet}, as well as more recent Transformer-based complex MTSF models \citep{wen2022transformers}. However, studies have pointed out that simpler univariate models, which ignore the relationships between series, can surpass these complex models \citep{zeng2022transformers,nie2023time}. This indicates that previous complex models may not effectively model these relationships.

For MTSF, effectively handling both temporal and contextual dimensions is crucial. Nonetheless, recent complex models are typically driven by 1D temporal attention mechanisms, as seen in LogTrans \citep{li2019enhancing}, Informer \citep{zhou2021informer}, Autoformer \citep{wu2022autoformer}, etc, which has been shown to inadequately capture series-wise relationships \citep{zeng2022transformers,lu2023arm}. Some research also tried to explicitly model series-wise relationships \citep{liu2023itransformer}, but the input temporal embedding diminish the effectiveness with longer input length. Crossformer \citep{zhang2023crossformer} tried to build a 2D attention mechanism, but its performance did not surpass linear univariate models. ARM \citep{lu2023arm} attempted to simulate 2D attention by weighting the results of temporal attention with channel attention scores, but its complexity and reliance on univariate components limit its effectiveness.

\section{Method}

In MTSF, our primary objective is to forecast the future values of a time series, represented as $\widehat{X}_P \in \mathbb{R}^{L_P \times C}$, given its historical data, denoted as $X_I \in \mathbb{R}^{L_I \times C}$. Here, $X \in \mathbb{R}^{L \times C}$ signifies the entire time series, where $L = L_I + L_P$ represents the total length of the series, $L_I$ is the input or lookback window length, $L_P$ is the forecasting horizon, and $C$ indicates the number of input series or channels. In this setting, $X^{(t), i}$ denotes the value at the $t$-th time step of the $i$-th series, with $t \in \{1, \ldots, L\}$ and $i \in \{1, \ldots, C\}$.

Addressing the intricacies of MTSF, we introduce the \underline{C}onstructing \underline{A}uxiliary \underline{T}ime \underline{S}eries (CATS) approach. CATS innovatively utilizes Auxiliary Time Series (ATS) to represent the inter-series relationship apart from the temporal relationship in the Original Time Series (OTS) $X$. The ATS is denoted as $A \in \mathbb{R}^{L \times N}$, where $N$ is the total number of ATS. These ATS are generated by $M$ ATS constructors $\{F_1, \ldots, F_M\}$, with each constructor $F_m: \mathbb{R}^{L_I \times C} \rightarrow \mathbb{R}^{L_I \times n_m}$ responsible for creating $n_m$ ATS, such that $\sum_{m=1}^M n_m = N$. 

\begin{figure}[ht]
\begin{center}
\centerline{\includegraphics[trim=5pt 8pt 5pt 0pt, clip, width=0.7\columnwidth]{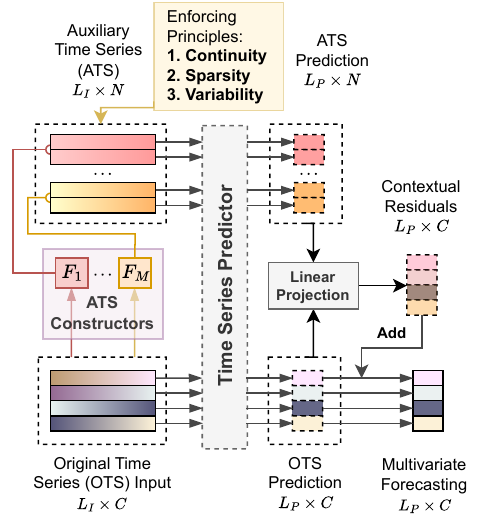}}
\caption{Overall Architecture of the CATS Approach}
\label{architecture}
\end{center}
\vskip -0.2in
\vspace{-12pt}
\end{figure}

\paragraph{CATS Approach}

The CATS structure, as shown in Figure \ref{architecture}, while conducting the prediction of the original series, constructs ATS capable of representing inter-series relationships and incorporate the ATS first-stage predicting results into the OTS first-stage prediction to obtain the model output. In the absence of any ATS constructor, the output of CATS is identical to the result obtained by directly using a time series predictor. This predictor can be a univariate or multivariate forecasting model with neural network architecture. The time series predictor is denoted as $\Phi_k: \mathbb{R}^{L_I \times k} \rightarrow \mathbb{R}^{L_P \times k}$, which is capable of mapping a specified $k$ number of time series inputs from input to forecasting horizon. The prediction process of the model without ATS is expressed as $\widehat{X}_P = \Phi_C(X_I)$. Let \([\ldots]\) denotes concatenation along the series channel dimension. With ATS $A$, the predicting process for the combined set of $N+C$ time series becomes $[\widehat{A}_P, \widehat{X}_P] = \Phi_{N+C}([A_I, X_I])$, where
$A_I = [F_{1}(X_I), F_{2}(X_I), \ldots, F_{M}(X_I)] \in \mathbb{R}^{L_P \times N}.$

The original $\widehat{X}_P$ part of the first-stage prediction is retained as an OTS shortcut, encompassing most of the intra-series temporal relationship. For the remaining $\widehat{A}_P$, concatenated with $\widehat{X}_P$, we then project the $N+C$ result channels at each timestep into the $C$-dimensional space of the OTS, yielding the $\widetilde{X}_P$ that primarily captures the multivariate relationship. The process can be expressed as: \(
\widetilde{X}_P^{(t), i} = \sum_{j=1}^{N+C} P_{ij} \cdot [\widehat{A}_P^{(t)},\widehat{X}_P^{(t)}]^j
\), where $\widetilde{X}_P$ denotes the integrated prediction and $P$ denotes the linear mapping matrix.
Finally, we can directly add the inter-series relationship component as the residual part to merge the first-stage prediction results as $\widehat{X}^*_P = \widehat{X}_P+\widetilde{X}_P$, where $\widehat{X}^*_P$ represents the final forecasting after correcting with multivariate information.


\subsection{Basic Intuition: How CATS Works}\label{intuition}
We present an intuitive example to demonstrate the efficacy and working mechanism of CATS. We apply the simplest form of ATS constructor -- identity mapping -- and use a basic independent linear predictor for first-stage prediction. Despite this ATS constructor merely replicates the OTS without any trainable parameter, it still remarkably transforms a univariate model into a model capable of capturing challenging multivariate relationships, through the mechanism similar to 2D temporal-contextual attention provided by CATS.

The example is a simple shifting problem introduced by \citep{lu2023arm} using a \textit{Multi} dataset, where the best solution is a copy-paste operation. However, most multivariate models struggle due to challenges in timestep mixing as well as managing relationship across different time steps of multivariate series (see the \textit{Multi} results in Table \ref{mainresults} and \ref{fullresults}). 

\textbf{Shifting Problem} \ \ Consider a scenario with two time series: a random walk series $X^1$, and another series $X^2$ derived by shifting $X^1$ backward by 96 steps. We set up a MTSF problem where the inputs comprises 144 steps (denoted as $X^1_I$ and $X^2_I$), and the goal is to predict the subsequent 48 steps (denoted as $X^1_P$ and $X^2_P$). For each series considered independently, the best univariate predictor $\widehat{X}^1_{Uni}$ and $\widehat{X}^2_{Uni}$ is a repetition of their last input value $X^{(144),1}_I$ and $X^{(144),2}_I$, respectively. However, when these series are considered together, an optimal predictor for $X^2$ would copy-paste steps 49 to 96 from $X^1$, denoted as $\widehat{X}^2_P=X^2_P=X^{(49:96),1}_I$.

\begin{figure}[ht]
\begin{center}
\centerline{\includegraphics[trim=0pt 3pt 0pt 0pt, clip, width=\columnwidth]{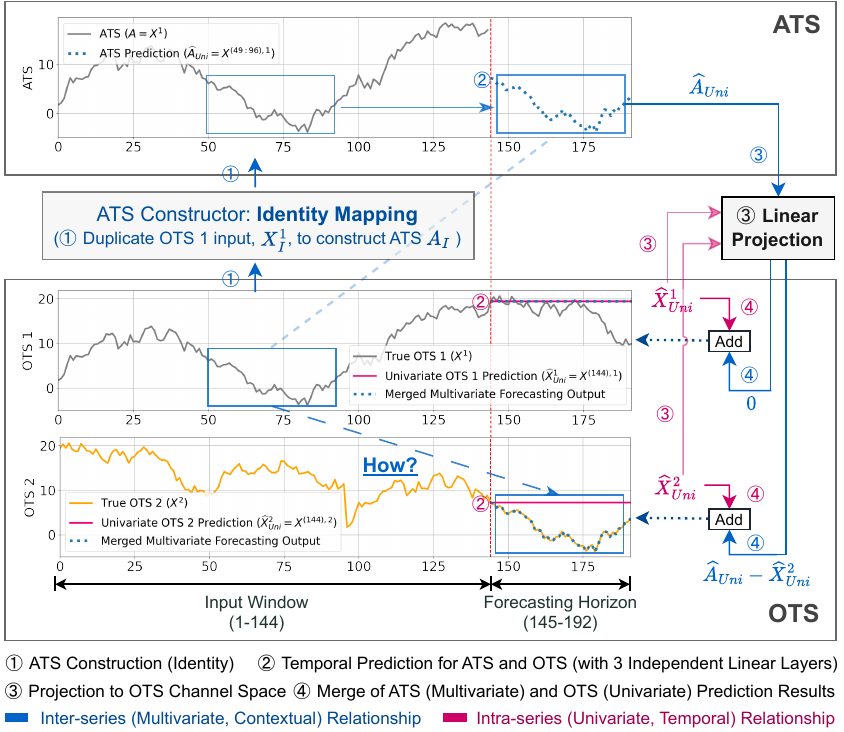}}
\caption{The Basic CATS Solution of the Shifting Problem}
\label{simpleintuition}
\end{center}
\vskip -0.2in
\end{figure}

Using the CATS structure, as shown in Figure \ref{simpleintuition}, we initially duplicate the first series to create an ATS, denoted as $A_I = X_I^1$. For the first-stage prediction, we use independent linear layers for each of $X^1$, $X^2$, and $A$. The best independent predictors for $X^1$ and $X^2$ will remain as their last values. However, the ATS's linear layer operates differently. Let the weight matrix for this layer be denoted by $W \in \mathbb{R}^{48 \times 144}$. When $W$ is structured such that its entries $W_{[:, 49:96]}=I$ and 0 otherwise, this layer effectively "transports" the steps 49 to 96 of the first series to the ATS's output with \( \widehat{A}_{Uni} = W A_I = X^{(49:96),1}_I\). Subsequently, a linear mapping then "transports" this segment from the ATS's output to the OTS output domain, making $\widehat{X}^2_P=X^{(49:96),1}_I$. 
In this simplified setting, even without complex network structure, CATS adaptively learns the inter-series relationship. This demonstrates CATS as a adaptive mechanism that focuses on specific regions of the OTS, summarize the inter-series information in ATS, and effectively transfers it from the input to the output domain, finally integrating it into the OTS output through projection.

\begin{figure}[ht]
\begin{center}
\centerline{\includegraphics[trim=0pt 3pt 0pt 0pt, clip, width=0.9\columnwidth]{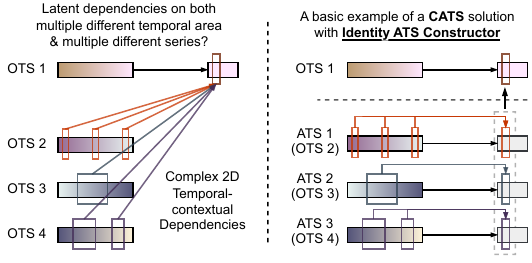}}
\caption{2D Temporal-contextual Attention-like Mechanism}
\label{tempcont}
\end{center}
\vskip -0.2in
\end{figure}

The shifting problem example shows that CATS can be viewed as an adaptive 2-dimensional temporal and series-wise attention mechanism: it dynamically focuses on specific regions of the OTS, and extract the inter-series information into ATS, effectively transfers this information from the input to the output time domain via the main predictor. Ultimately, it merges the gathered information into the OTS output using a projection, as illustrated in Figure \ref{tempcont}. For more complicated time series, CATS will expand the number of ATS and use general ATS constructors.

\subsection{Three Key Principles of ATS}

This section discusses three critical principles: continuity, sparsity, and variability. These principles substantially enhance both the performance and flexibility of the model, enabling ATS to adaptively and accurately extract useful information and represent inter-series dependencies. This facilitates correct pattern fitting across different datasets.

\subsubsection{Continuity}

Continuity is an important and inherent characteristic for most real-world time series datasets. We use a continuity loss term \( \mathcal{L}_{\text{cont}} \) to enforce this principle of continuity in ATS:
\[
\mathcal{L}_{\text{cont}} = \frac{\beta_{cont}}{L_I N} \sum_{t=2}^{L_I} \sum_{n=1}^{N} \left(\frac{A^{(t), n}-A^{(t-1), n}}{\sigma_{n}}\right)^2
\]
where \(\beta_{cont}\) is a weighting factor that adjusts the influence of the continuity loss in the overall loss function. \( \sigma_{n} \) is standard deviation of the ATS for channel \( n \). CATS is trained using a Mean Squared Error (MSE) loss combined with the continuity loss. By applying continuity loss, ATS will smooth out temporal details and noise, focusing on overall trends that can be described by inter-series dependencies, leaving the remaining intra-series details to be better modeled by established methods on the OTS. This clear division of modeling focus using ATS and OTS enhances the reliability and reduces the complexity of forecasting. We further investigate this from two theoretical perspectives. Detailed explanation of this part can be found in Appendix \ref{additionalcont}.




\textbf{Spectral Density and Filtering} \ \  The introduction of continuity loss in ATS can be regarded as applying a low-pass filter. This filter effectively diminishes the high-frequency components in the ATS, making it smoother and less sensitive to noise. This can be expressed as a modification of the spectral density \( S_X(f) \) of $X$, where \( S_A(f) = S_X(f) \cdot H(f) \), with \( H(f) \) being the low-pass filter. Consequently, the ATS focuses more on the overall trends rather than short-term fluctuations, providing a clearer representation of inter-series relationships.

\textbf{Quadratic Variation and Denoising} \ \   From the perspective of quadratic variation, continuity in the CATS model aids in reducing the ATS volatility. For a Brownian motion-driven process \( X(t) \), the quadratic variation is significant. However, for ATS \( A(t) \) with its quadratic variation close to zero, it will show more properties like a continuous differentiable function. This reduction in quadratic variation indicates that ATS with continuity becomes smoother and less prone to fluctuations like white noise, thereby enhancing the ability to capture and represent more long-term trends in the OTS.
 
Therefore, with continuity loss, the ATS first-stage prediction component \( \widetilde{X}_P \) will primarily extract inter-series trends, allowing the OTS prediction component \( \widehat{X}_P \) to focus on the more stable intra-series autocorrelation. This aligns perfectly with the structure and objectives of CATS.

\subsubsection{Sparsity}
\textbf{Channel Sparsity} \ \ In CATS, ATS are tasked with representing inter-series dependencies. Therefore, when there are more ATS channels, they tend to capture more inter-series information to influence the final forecasting. For many datasets where inter-series dependencies are weak, forcibly seeking them can lead to significant overfitting. Ideally, the number of available ATS should adaptively decrease when dealing with datasets that have weaker multivariate relationships. By activating only the most crucial ATS channels, model stability and performance would significantly improve. It also reduces the tuning cost for the hyperparameter of the number of ATS series. We refer to this as the principle of channel sparsity.

\begin{figure}[ht]
\begin{center}
\centerline{\includegraphics[width=0.90\columnwidth]{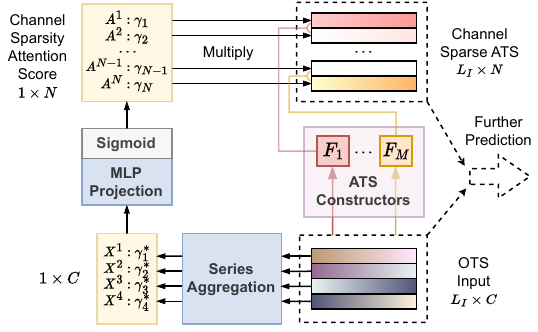}}
\caption{Channel Attention Score Calculation of Channel Sparsity}
\label{channelsparsity}
\end{center}
\vskip -0.2in
\end{figure}

A mechanism is designed to allow the model to dynamically activate a varying number of ATS based on the strength of inter-series relationships, as shown in Figure \ref{channelsparsity}. Similar to the channel attention mechanism commonly used in computer vision \citep{wang2017residual, Hu_2018_CVPR}, this method aggregates the input OTS channels and conducts upscaling projection to obtain attention scores for each ATS channel. Ideally, the channel sparsity attention scores should be mostly zeros. However, to prevent the ATS from being insufficiently trained due to gradient problems, we use soft attention scores instead of hard channel dropping here.

For the OTS input \( X_I \in \mathbb{R}^{L_I \times C} \), we first aggregate each series \( i \) through a linear layer to obtain the corresponding scalar \( \gamma^*_i \). which can be expressed as
\(
\gamma^*_i = \mathbf{w}^\top X^i + b
\),
where \( \mathbf{w} \in \mathbb{R}^{L_I} \) and \( b \in \mathbb{R} \) are the weight and bias of the linear layer. Then, we combine all \( \gamma^*_i \) into a vector and pass it through a upscaling MLP with activation to obtain the attention score \( \gamma_n \) for each ATS, which can be described as:
\[
\boldsymbol{\gamma} = \mathtt{Sigmoid}\left(W_2 \cdot \mathtt{GELU}\left(W_1 \boldsymbol{\gamma}^* + \mathbf{b}_1\right) + \mathbf{b}_2\right)
\]
where \( \boldsymbol{\gamma}^* = [\gamma^*_1, \gamma^*_2, \ldots, \gamma^*_C]^\top \), \( W_1 \in \mathbb{R}^{U \times C} \) and \( W_2 \in \mathbb{R}^{N \times U} \) are the weights of the MLP, \( \mathbf{b}_1 \in \mathbb{R}^{U} \) and \( \mathbf{b}_2 \in \mathbb{R}^{N} \) are the biases of the MLP, and \( U \) is the dimension of the hidden layer. The applying of \( \boldsymbol{\gamma} \) on \( A_I \) can be represented as: \(A'_I = A_I \odot \boldsymbol{\gamma}\), where \( A'_I \) is the adjusted ATS after applying the attention scores \( \boldsymbol{\gamma} \), \( \odot \) denotes element-wise multiplication. \( \boldsymbol{\gamma} \) are used to suppress those unimportant ATS channels, thus allowing the strength of the inter-series relationships represented by ATS to dynamically self-adjust.

\textbf{Temporal Sparsity} \ \ Previous MTSF research has shown that altering the length $L_I$ significantly impacts model performance, leading most models to conduct hyper-parameter tuning of $L_I$ \citep{wu2022autoformer, zhou2022fedformer, zeng2022transformers, nie2023time, lu2023arm}. Choosing a too short $L_I$ may fail to learn the correct dependencies, while a too long $L_I$ can result in overfitting. 
Clearly, a fixed $L_I$ for all the series is not the optimal choice. We need a mechanism to adaptively cutoff the time series input of the predictor. In this scenario, each channel can discard different lengths of long-term parts that are unnecessary for its prediction. Hence, the predictor input will include time series of varying lengths. We refer to this principle as temporal sparsity.


\begin{figure}[ht]
\begin{center}
\centerline{\includegraphics[width=0.95\columnwidth]{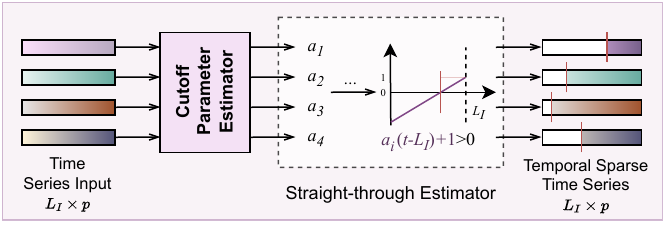}}
\caption{Adaptive Temporal Cutoff of Temporal Sparsity}
\label{temporalsparsity}
\end{center}
\vskip -0.2in
\end{figure}

An adaptive temporal cutoff method is designed to determine the required cutoff length for each series, as shown in Figure \ref{temporalsparsity}. Let the input series at this stage be $Q_I \in \mathbb{Q}^{L_I \times p}$ with $p$ channels. We use a parameter $a_i$ to determine the cutoff length for each series $Q^i$. For every $Q^i$, due to potential differences in properties, we assign independent linear layers to calculate their $a_i= \mathbf{W}^i Q_I^i + \mathsf{b}_i$, where \( \mathbf{W} \in \mathbb{R}^{L_I \times p} \) and \(\mathsf{b} \in \mathbb{R}^{p}\) are weights and biases. Then, we construct a linear function $h_i(t)$ as our hard cutoff function, where $t$ is each timestep. We calculate $h_i(t) = a_i(t - L_I) + 1$, and determine whether $h_i(t) > 0$ is true for each step $t$ to obtain the cutoff result \( Q'^{(t),i} = Q^{(t),i} \cdot \mathbf{1}_{\{h_i(t) > 0\}} \), where \( \mathbf{1}_{\{\cdot\}} \) is the indicator function. If the inequality is not true, we set the value at $t$ to 0, thus achieving the purpose of cutting off the earlier part of the series. However, this indicator function is not differentiable and may disrupt the learning. Therefore, we use a straight-through estimator, copying the gradient of $Q'^{(t),i}$ to $Q^{(t),i}$ to bypass this non-differentiable part. Then, $Q'^{(t),i}$ becomes:
\[
Q'^{(t)}_i = Q^{(t)}_i \cdot \left(h_i(t) - \mathtt{sg}\left(h_i(t) - \mathbf{1}_{\{h_i(t) > 0\})} \right)\right)
\]
where $\mathtt{sg}(\cdot)$ denotes the stop-gradient operation. Through this method, we obtain a time series with temporal sparsity after dynamic truncation. In the implementation, we apply the temporal sparsity module on both the ATS and OTS.

\subsubsection{Variability}
Variability represents the diversity of inter-series dependencies that ATS can capture, reflecting how effectively OTS information is combined in varied ATS from multiple perspectives. This approach helps to prevent the forecasting from being controlled by a specific type of ATS, allowing the model to adapt to diverse MTSF datasets with different characteristics of dependency. CATS achieves this through a variety of ATS constructors with differing structures.

\subsection{ATS Constructors}\label{constructorsection}
\begin{figure}[t]
\vskip 0.05in
\begin{center}
\centerline{\includegraphics[width=0.9\columnwidth]{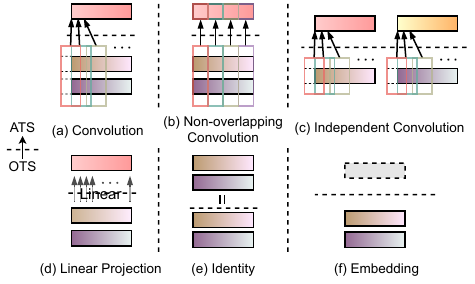}}
\caption{Different Types of ATS Constructors}
\label{constructors}
\end{center}
\vskip -0.3in

\end{figure}

We develop a series of ATS constructors \( F_m \) as shown in Figure \ref{constructorsection}. With the benefits of CATS, we can simultaneously use multiple different structures as feature extractors. This allows us to surpass the constraints of previous MTSF models, which often restrict themselves to specific types of temporal processing methods. Note that we use GELU activation after each \( F_m \) to add non-linearity. The settings of ATS constructors used in our experiments are detailed in Appendix \ref{hyperparametersettings} and Table \ref{additionalhyperparameter}.

\textbf{Convolution} is a prevalent temporal feature extractor that is capable of learning the contribution of local patterns to forecasting. The impact of varying kernel sizes in convolution is well-documented. \citep{lai2018modeling, borovykh2017conditional, vanwavenet, li2019enhancing, lu2023arm}. CATS employs convolution layers with different kernel sizes as ATS constructors to accommodate different scenarios. The $m$th constructor utilizes 1D convolution $\mathtt{Conv1d}_{[C \rightarrow n_m]}(\cdot, K, D, S, G)$, defined by kernel size \(K\), padding \(D\), stride \(S\), channel groups \(G\), and the \(n_m\) output channels mapped from \(C\) OTS. The constructor is given by: 
\(F^{[\mathtt{Conv}]}_m (X_I) = \mathtt{Conv1d}_{[C \rightarrow n_m]}(X_I^{\top}, K, \lfloor \frac{K-1}{2} \rfloor, 1, 1)^{\top}\).

\textbf{Non-overlapping Convolution}, similar to patching, temporal aggregation, or hierarchical grouping, uses a convolution layer with kernel size equal to stride, which is widely applied in computer vision and time series \citep{dosovitskiy2021an, liu2021swin, nie2023time}.  Since we should keep the output length as \( L_I \), we need some extra padding $E=(K-(L_I \ \ \mathtt{mod} \ \ K)) \ \ \mathtt{mod} \ \ K$ and for each input-to-output patch, do $K$ times of Conv1d calculation:
\(F^{[\mathtt{NOConv}]}_m (X_I) = \mathtt{Conv1d}_{[C \rightarrow n_m]}(X_I^{\top}, K, \lfloor \frac{E}{2} \rfloor, K, 1)^{\top}\).

\textbf{Independent Convolution} performs 1D convolution on each series separately. The ability of independent convolution to capture effective univariate features has been widely validated in previous research \citep{lai2018modeling, borovykh2017conditional}. By extracting local features independently for each series, it enhances ATS's capability to recognize important series and avoids timestep mixing errors. The number of output channels could be $n_m=v \cdot C$, where $v \in \mathbb{Z}^+$. The constructor can be expressed as:
\(F^{[\mathtt{IConv}]}_m (X_I) = \mathtt{Conv1d}_{[C \rightarrow v \cdot C]}(X_I^{\top}, K, \lfloor \frac{K-1}{2} \rfloor, 1, C)^{\top}\).

\textbf{Linear Projection} This fundamental feature extractor linearly maps OTS across channels to form latent auxiliary series. Frequently used in MTSF models like Transformers and MLP-Mixers for input processing \citep{zhou2021informer, wu2022autoformer, zhang2023crossformer}, linear projection offers a straightforward dimensional transformation. However, it can mix unrelated series, a risk CATS mitigates by combining various ATS constructors and retaining an OTS shortcut. This constructor is expressed as: 
\( F^{[\mathtt{Lin}]}_m(X_I) = \mathtt{Conv1d}_{[C \rightarrow n_m]}(X_I^{\top}, 1, 0, 0, 1)^{\top} \)


\textbf{Identity} Proven effective in \S \ref{intuition}, this method retains the input series' original information, facilitating efficient learning of fundamental inter-series relationships like mapping and shifting. The constructor is simply: \(F^{[\mathtt{Id}]}_m(X_I)=X_I\).

\textbf{Embedding} involves a trainable matrix unrelated to OTS values but linked to timesteps. Similar to positional embeddings in Transformers, CATS adds embedding channels with time series properties, aiding in understanding temporal connections. The constructor, with a learnable matrix \( \mathcal{W}^{[\mathtt{Emb}]} \in \mathbb{R}^{n_m \times C} \), is: \(F^{[\mathtt{Emb}]}_m(X_I)=\mathcal{W}^{[\mathtt{Emb}]}\).

\subsection{Time Series Predictor}
The enough inductive bias provided by CATS allows us to choose a simple main predictor \( \Phi_{N+C} \), whose primary task shifts to mapping the effective information from the input domain to the correct positions in the forecasting horizon. Thus, in the basic implementation, we simply use a two-layer MLP (2L) as the main predictor. Other structures, such as previous MTSF models (e.g., DLinear, Autoformer), independent linear layers (IndLin), and even the simplest average pooling (Mean), are also used in our ablation studies \S \ref{ablation}. For details on the structure of the predictors, see \S \ref{timeseriespredictor}.

\section{Experiments and Results}

Comprehensive experiments were conducted on nine commonly-used MTSF datasets (details in Appendix \ref{datasource}), including ETT \citep{zhou2021informer}, Traffic, Electricity, Weather, Exchange Rate \citep{lai2018modeling}, and Multi \citep{lu2023arm}. We compared CATS against baseline models from six recent MTSF studies: ARM \citep{lu2023arm}, PatchTST \citep{nie2023time}, DLinear \citep{zeng2022transformers}, FedFormer \citep{zhou2022fedformer}, Autoformer \citep{wu2022autoformer}, and Informer \citep{zhou2021informer}. We use the same experiment environment as these models for fair comparison. In CATS, a structurally simple two-layer MLP (2L) was employed as the time series predictor to demonstrate CATS's capability of empowering univariate models to effectively learn inter-series relationships. We use a total of 8 ATS constructors, detailed in Table \ref{additionalhyperparameter}. We trained CATS with a combination of Mean Squared Error (MSE) and continuity loss, using the Adam optimizer \citep{KingBa15}. Also, we applied common training strategies from recent studies, including last-value demeaning and random dropping \citep{kim2021reversible, zeng2022transformers, lu2023arm}. To identify the enhancement specific from CATS, these strategies were also applied across all baseline models in subsequent ablation studies (\S \ref{ablation}). Models from Informer to PatchTST required tuning of the input length $L_I$, selecting from $L_I \in \{96, 192, 336, 720\}$ for optimal results. However, CATS's temporal sparsity module enables dynamic input cutoff for each channel, allowing a fixed $L_I = 720$. In the main text, the performance metrics used included averaged MSE and calculated average MSE percentage compared to a pivotal model on the test set. For original MSE and Mean Absolute Error (MAE) data, please refer to the Appendix \ref{addtionalresults}. Details on hyper-parameter settings and implementations can be found in Appendix \ref{implementationdetails}. Some visualization examples of ATS and OTS results can be found in Figure \ref{visualizationATS}.

\textbf{Main MTSF Experiment Results} \ \ Table \ref{mainresults} highlights the superior performance of CATS across widely-used real-world datasets, outperforming existing MTSF models in average MSE and achieving a average ranking close to the first place. In 28 out of 36 sub-experiments for different $L_P$ , CATS consistently delivered best outcomes (see Table \ref{fullresults}). This reflects the significant enhancement brought by constructed ATS, thereby empowering the a simple univariate 2-layer MLP with temporal-contextual modeling capacity.

\begin{table}[tb]
\renewcommand{\arraystretch}{0.9}
\caption{Summary of MTSF results with forecasting horizons \( L_P \in \{96,192,336,720\} \). \textbf{See \underline{Table \ref{fullresults}}} for original results. Averages of MSE results (in parentheses) for each model on each dataset and MSE percentages (above parentheses) relative to Repeat are provided. The best and second best results are in bold and underlined, respectively. Average percentages (Avg\%) across all datasets and average rankings (AvgRank) of each model, along with the count of first-place rankings (\#Win), are also included.}
\label{mainresults}
\centering
\begin{threeparttable}
\begin{small}
\setlength{\tabcolsep}{1pt}
\resizebox{\columnwidth}{!}{
\begin{tabular}{lccccccc}
\toprule
 
Models &
\textbf{CATS (2L)} &
ARM &
PatchTST &
DLinear &
Autoformer &
Informer & 
Repeat
\\
\midrule
\noalign{\vskip 1pt}
\multirow{2}{*}{Electricity} & \textbf{9.24\%} & \underline{9.31\%} & 9.87\% & 10.3\% & 14.1\% & 19.3\% & 100.0\% \\
& (\textbf{0.149}) & (\underline{0.150}) & (0.159) & (0.166) & (0.227) & (0.311) & (1.612) \\
\noalign{\vskip 1pt}
\multirow{2}{*}{ETTm1}  & \textbf{27.1\%} & \underline{27.4\%} & 27.8\% & 28.1\% & 40.6\% & 68.7\% & 100.0\% \\
& (\textbf{0.345}) & (\underline{0.348}) & (0.353) & (0.357) & (0.515) & (0.872) & (1.269) \\
\noalign{\vskip 1pt}
\multirow{2}{*}{ETTm2}  & \textbf{63.2\%} & \underline{65.2\%} & 66.6\% & 69.5\% & 84.3\% & 366.5\% & 100.0\% \\
& (\textbf{0.243}) & (\underline{0.251}) & (0.256) & (0.267) & (0.324) & (1.410) & (0.385) \\
\noalign{\vskip 1pt}
\multirow{2}{*}{ETTh1} & \underline{30.9\%} & \textbf{30.8\%} & 31.3\% & 32.0\% & 35.8\% & 78.2\% & 100.0\% \\
& (\underline{0.408}) & (\textbf{0.407}) & (0.413) & (0.423) & (0.473) & (1.033) & (1.321) \\
\noalign{\vskip 1pt}
\multirow{2}{*}{ETTh2}  & \textbf{58.5\%} & \underline{61.5\%} & 61.7\% & 80.4\% & 78.7\% & 615.9\% & 100.0\% \\
& (\textbf{0.314}) & (\underline{0.330}) & (0.331) & (0.431) & (0.422) & (3.303) & (0.536) \\
\noalign{\vskip 1pt}
\multirow{2}{*}{Weather}  & \textbf{59.5\%} & \underline{61.1\%} & 64.0\% & 69.8\% & 95.8\% & 179.8\% & 100.0\% \\
& (\textbf{0.210}) & (\underline{0.215}) & (0.226) & (0.246) & (0.338) & (0.634) & (0.353) \\
\noalign{\vskip 1pt}
\multirow{2}{*}{Traffic}  & \textbf{13.7\%} & \underline{13.9\%} & 14.1\% & 15.7\% & 22.7\% & 27.6\% & 100.0\% \\
& (\textbf{0.379}) & (\underline{0.384}) & (0.391) & (0.434) & (0.628) & (0.764) & (2.770) \\
\noalign{\vskip 1pt}
\multirow{2}{*}{Exchange}  & \underline{73.0\%} & \textbf{70.2\%} & 109.0\% & 86.2\% & 178.3\% & 450.7\% & 100.0\% \\
& (\underline{0.251}) & (\textbf{0.242}) & (0.375) & (0.297) & (0.613) & (1.550) & (0.344) \\
\noalign{\vskip 1pt}
\multirow{2}{*}{Multi}  & \textbf{59.6\%} & \underline{64.9\%} & 114.7\% & 84.1\% & 164.4\% & 137.7\% & 100.0\% \\
& (\textbf{0.168}) & (\underline{0.177}) & (0.313) & (0.230) & (0.449) & (0.376) & (0.273) \\
\noalign{\vskip 1pt}

\midrule 

\textbf{Avg\%} & \textbf{43.9\%} & \underline{44.9\%} & 55.4\% & 52.9\% & 79.4\% & 216.0\% & 100.0\% \\
AvgRank & \textbf{1.25} & \underline{1.75} & 3.47 & 3.83 & 6.28 & 7.39 & 6.69 \\
\#Win   & \textbf{28} & \underline{10} & 1 & 0 & 0 & 0 & 0 \\

\bottomrule
\end{tabular}
}
\end{small}
\end{threeparttable}
\vspace{-12pt}
\end{table}

\textbf{Extended MTSF Experiment Results} \ \ 
In \S \ref{intuition}, we showed a shifting problem, which is introduced by ARM \citep{lu2023arm} to create the Multi dataset in our benchmarks to test models in strong multivariate relationship scenarios, where CATS demonstrated superior performance. To further assess CATS in more complex conditions with a greater number of series than 8 in Multi, we created the MultiX datasets. These datasets derive each series from a master AR(1) series, with each subsequent series lagging the previous by a set stride, adding and accumulating random noise in each shifting to challenge the model's ability to learn generative dependencies rather than just relying on the first series. Eventually, the final series is noised offset from the first by a total of $S$ steps. We set $S=L_I=720$. If the number of extracted series, X, is 100, the dataset is named Multi100, and so forth. More details about this dataset are in Appendix \ref{datasource}.

\begin{table}[tb]
\renewcommand{\arraystretch}{0.9}
\caption{Summary of extended MTSF results on datasets with strong inter-series relationship. \textbf{See \underline{Table \ref{fullmultiresults}}} for original results. The method of displaying summary data is the same as in Table \ref{mainresults}.}
\label{multiresult}
\centering
\begin{threeparttable}
\begin{small}
\setlength{\tabcolsep}{1pt}
\resizebox{\columnwidth}{!}{
\begin{tabular}{lccccccc}
\toprule
 
Models &
\textbf{CATS (2L)} &
ARM &
PatchTST &
DLinear &
Autoformer &
Informer & 
Repeat
\\
\midrule
\noalign{\vskip 1pt}
\multirow{2}{*}{Multi20}  & \textbf{58.2\%} & \underline{64.2\%} & 95.3\% & 86.8\% & 200.0\% & 293.8\% & 100.0\% \\
& (\textbf{0.017}) & (\underline{0.019}) & (0.027) & (0.025) & (0.058) & (0.085) & (0.029)\\
\noalign{\vskip 1pt}
\multirow{2}{*}{Multi50}  & \textbf{62.9\%} & \underline{70.6\%} & 95.9\% & 85.4\% & 249.2\% & 356.6\% & 100.0\% \\
& (\textbf{0.018}) & (\underline{0.021}) & (0.028) & (0.025) & (0.072) & (0.104) & (0.029)\\
\noalign{\vskip 1pt}
\multirow{2}{*}{Multi100}  & \textbf{61.8\%} & \underline{69.5\%} & 96.0\% & 86.2\% & 409.4\% & 384.2\% & 100.0\%\\
& (\textbf{0.018}) & (\underline{0.020}) & (0.028) & (0.025) & (0.119) & (0.112) & (0.029)\\
\noalign{\vskip 1pt}
\multirow{2}{*}{Multi200}  & \textbf{55.7\%} & \underline{77.7\%} & 92.3\% & 86.9\% & 243.2\% & 275.8\% & 100.0\%\\
& (\textbf{0.017}) & (\underline{0.023}) & (0.027) & (0.026) & (0.072) & (0.082) & (0.030)\\
\noalign{\vskip 1pt}

\midrule 

\textbf{Avg\%} & \textbf{59.6\%} & \underline{70.5\%} & 94.9\% & 86.3\% & 275.5\% & 327.6\% & 100.0\%\\
AvgRank & \textbf{1.50} & \underline{1.94} & 4.06 & 2.81 & 7.06 & 7.06 & 4.75 \\
\#Win   & \textbf{12} & 1 & 0 & \underline{3} & 0 & 0 & 0\\

\bottomrule
\end{tabular}
}
\end{small}
\end{threeparttable}
\vspace{-12pt}
\end{table}

Table \ref{multiresult} shows CATS's exceptional performance in MultiX scenarios. CATS significantly outperforms previous models in average MSE and achieves the best results in 12 out of 16 sub-experiments. 
As the forecasting horizon expands, the influence of true inter-series relationships diminishes, leading to a scenario resembling a univariate AR(1) prediction with noise. In such cases, CATS's channel sparsity feature becomes more prominent, effectively reducing the ATS's influence and reverting the model to a simpler form of a univariate MLP. This behavior demonstrates CATS's adaptability in handling varying strengths of multivariate relationships. We further discuss the importance of using MultiX as a benchmark dataset in \S \ref{multixdiscussion}.

\begin{table*}[tb]
\renewcommand{\arraystretch}{0.8}
\caption{Summary of results for using CATS with time series predictors. \textbf{See \underline{Table \ref{fullstructureresults}}} for original results. We compared the average MSE before and after using CATS and calculated the percentage based on the averages without CATS. Better results are highlighted in bold.}
\label{ablationcats}
\centering
\begin{threeparttable}
\begin{small}
\setlength{\tabcolsep}{12pt}
\resizebox{\textwidth}{!}{
\begin{tabular}{l|cc|cc|cc|cc|cc}
\toprule
 
\multirow{2}{*}{Models} &
\multirow{2}{*}{2L} &
2L &
\multirow{2}{*}{DLinear} &
DLinear &
\multirow{2}{*}{Autoformer} &
Autoformer &
\multirow{2}{*}{IndLin} &
IndLin &
\multirow{2}{*}{Mean} &
Mean
\\
& & +CATS & & +CATS & & +CATS & & + CATS & & + CATS
\\
\cmidrule(lr){1-1}\cmidrule(lr){2-3}\cmidrule(lr){4-5}\cmidrule(lr){6-7}\cmidrule(lr){8-9}\cmidrule(lr){10-11}

\multirow{2}{*}{Multi}  & 100.0\%  & \textbf{45.2\%}  & 100.0\% & \textbf{63.2\%} & 100.0\% & \textbf{36.0\%} & 100.0\% & \textbf{57.7\%} & 100.0\% & \textbf{37.9\%} \\
& (0.360) & (\textbf{0.163}) & (0.286) & (\textbf{0.181}) & (0.595) & (\textbf{0.214}) & (0.291) & (\textbf{0.168}) & (0.654) & (\textbf{0.248}) \\
\noalign{\vskip 1pt}
\multirow{2}{*}{ETTh2} & 100.0\% & \textbf{90.3\%} & 100.0\% & \textbf{98.1\%} & 100.0\% & \textbf{89.8\%} & 100.0\% & \textbf{98.8\%} & 100.0\% & \textbf{92.6\%} \\
& (0.348) & (\textbf{0.314}) & (0.341) & (\textbf{0.335}) & (0.390) & (\textbf{0.351}) & (0.340) & (\textbf{0.336}) & (0.408) & (\textbf{0.378}) \\

\bottomrule
\end{tabular}
}
\end{small}
\end{threeparttable}
\vspace{-12pt}
\end{table*}

\subsection{Ablation Studies}\label{ablation}
In the ablation studies, we apply CATS to basic time series predictors and previous MTSF models to demonstrate the enhancement attributed to the CATS structure. We also show the effects of the three key principles of ATS. For the experiments, we select two datasets, Multi and ETTh2, which respectively represent scenarios with strong and weak inter-series dependencies. To eliminate the impact of training strategies, including last-value demeaning and random dropping, we implement these strategies across all baseline models during training. We fix the input length \(L_I = 720\) and only alter the elements of CATS components described in Figure \ref{architecture}, while keeping all other settings constant.

\begin{table*}[tb]
\renewcommand{\arraystretch}{0.8}
\caption{Summary of the ablation studies of the three ATS principles and OTS shortcut. \textbf{See \underline{Table \ref{fullablationresult}}} for original results. We calculated the average MSE and percentages based on the full CATS model to show the impact of these components on the model performance.}
\label{ablationproperties}
\centering
\begin{threeparttable}
\begin{small}
\setlength{\tabcolsep}{11pt}
\resizebox{\textwidth}{!}{
\begin{tabular}{lcccccccc}
\toprule
 
\multirow{2}{*}{Models} &
\multirow{2}{*}{\textbf{CATS (2L)}} &
\multirow{2}{*}{w/o Continuity} &
w/o Sparsity &
w/o Sparsity &
w/o Sparsity &
w/o Variability &
w/o Variability &
w/o OTS
\\
& & & (Channel) & (Temporal) & (Both) & (Pure Linear) & (Pure Conv) & Shortcut
\\
\midrule
\multirow{2}{*}{Multi} & \textbf{100.0\%} & 141.9\% & 156.7\% & 126.4\% & 158.4\% & 146.1\% & 122.1\% & 128.3\%\\
& (\textbf{0.163}) & (0.231) & (0.255) & (0.206) & (0.258) & (0.238) & (0.199) & (0.209)\\
\noalign{\vskip 1pt}
\multirow{2}{*}{ETTh2}  & \textbf{100.0\%} & 103.5\% & 103.2\% & 104.1\% & 105.3\% & 104.1\% & 104.2\% & 109.6\%\\
& (\textbf{0.314}) & (0.325) & (0.324) & (0.327) & (0.330) & (0.327) & (0.327) & (0.344)\\

\bottomrule
\end{tabular}
}
\end{small}
\end{threeparttable}
\vspace{-12pt}
\end{table*}

\textbf{Time Series Predictors + CATS} \ \ 
In Table \ref{ablationcats}, we apply the CATS structure to a univariate model (DLinear), a multivariate model (Autoformer), and three basic univariate predictors (2L, IndLin, Mean). IndLin represents the use of independent linear layers for each ATS/OTS first-stage prediction. Mean fills the corresponding \(L_P\) region for each ATS/OTS with its mean value of the input.

The integration of CATS significantly enhances the performance of each baseline. CATS provides significant and stable performance enhancements in scenarios with strong multivariate relationships and effectively suppresses overfitting when these relationships are weaker, ensuring better results than the baselines. These enhancements are independent of the predictor's structure. Even when the predictor is merely performing an average pooling (Mean), CATS is still able to transfer the inter-series relationships into the OTS output, surpassing the performance of most baselines. Note that IndLin performs the best in the sub-experiments with very strong inter-series dependencies, as shown in Table \ref{fullstructureresults}, which aligns with the example in \S \ref{intuition}. This highlights the powerful and flexible foundational fitting ability provided by the 2D attention-like structure for multivariate modeling.

\textbf{Three ATS Principles} \ \ 
Table \ref{ablationproperties} investigates the influence of key principles on performance. It shows that the absence of any principles significantly diminishes performance.

Continuity allows the ATS to focus more on inter-series trend effects, enabling the OTS part to better model the stabler autocorrelation after removing the inter-series long-term trend. This enhancement is particularly notable in ETTh2, characterized by many long-term trend changes.

Regarding sparsity, removing both sparsity modules drastically reduces performance. In the Multi dataset, where we aim to model numerous inter-series relationships using a sufficient number of ATS series, selecting the important series is vital to prevent overfitting. In ETTh2, characterized by weaker multivariate effects, using channel sparsity to manage ATS strength significantly enhances outcomes. Also,adjusting \(L_I\) has been shown to affect ETTh2's results markedly \citep{nie2023time,zeng2022transformers}. Implementing temporal sparsity for dynamic lookback further improves performance.

Variability influences the scope of inter-series dependencies captured by ATS. We tested this by altering ATS constructors: first, all constructors were changed to linear projections (Pure Linear), leading to reduced performance on both datasets. Next, using solely 1D convolutions (Pure Conv) also fell short compared to a mix of constructors, but slightly outperformed Pure Linear on the Multi dataset, highlighting the importance of local and series-independent feature processing for stronger inter-series relationship representation.

We assessed the effect of the OTS shortcut on performance. This feature, alongside sparsity, guarantees that CATS's results are at least as good as the base model. In ETTh2, characterized by weaker inter-series and stronger intra-series dependencies, removing the OTS shortcut led to significant performance drops due to the loss of series-wise independence, leading to improper mixing of series information in ATS-to-OTS projections. This highlights how the integration of various CATS structural components adaptively enhances performance.

\textbf{Computational Costs} \ \ 
As illustrated in Table \ref{computationalcostresult}, compared to Transformer-based models like ARM, Autoformer, Informer, and PatchTST, CATS significantly reduces both computational complexity and parameter size. When compared with simple univariate models such as DLinear, CATS effectively models varying strengths of inter-series dependencies with an acceptable incremental computational cost. Consequently, CATS can be characterized as a lightweight, low-complexity, and easily transferable structure for MTSF.

\begin{table}[h]
\renewcommand{\arraystretch}{0.8}
  \caption{Comparison of computational costs. This comparison utilizes the data format of ETTh2 to construct model inputs. $L_I$ is set to 720, and the other hyper-parameters for every model are set according to their default configurations.}\label{computationalcostresult}
  \centering
  \begin{threeparttable}
  \begin{small}
  \renewcommand{\multirowsetup}{\centering}
  \setlength{\tabcolsep}{7pt}
  \resizebox{\columnwidth}{!}{ 
\begin{tabular}{lcccccc}
\midrule
 Models & \multicolumn{2}{c}{\textbf{CATS (2L)}} & \multicolumn{2}{c}{ARM} & \multicolumn{2}{c}{Autoformer} \\
 \cmidrule(lr){2-3} \cmidrule(lr){4-5}\cmidrule(lr){6-7}
 Metric & FLOPs & Params & FLOPs & Params & FLOPs & Params \\
\midrule
$L_P=96$ & 382M & 2.47M & 426M & 7.89M & 10.9G & 15.5M \\
$L_P=192$ & 422M & 2.74M & 515M & 10.4M & 11.6G & 15.5M \\
$L_P=336$ & 483M & 3.16M & 664M & 14.9M & 12.7G & 15.5M \\
$L_P=720$ & 645M & 4.26M & 1.15G & 30.6M & 15.5G & 15.5M \\
\midrule
Models & \multicolumn{2}{c}{Informer} & \multicolumn{2}{c}{PatchTST} & \multicolumn{2}{c}{DLinear} \\
 \cmidrule(lr){2-3} \cmidrule(lr){4-5}\cmidrule(lr){6-7}
 Metric & FLOPs & Params & FLOPs & Params & FLOPs & Params \\
\midrule
$L_P=96$ & 9.41G & 11.3M & 5.26G & 4.87M & 3.06M & 138K \\
$L_P=192$ & 10.1G & 11.3M & 5.31G & 7.08M & 6.10M & 277K \\
$L_P=336$ & 11.2G & 11.3M & 5.38G & 10.4M & 10.7M & 485K \\
$L_P=720$ & 14.0G & 11.3M & 5.58G & 19.2M & 22.8M & 1.04M \\
\bottomrule
\end{tabular}
  }
  \end{small}
  \end{threeparttable}
\vspace{-12pt}
\end{table}

\section{Conclusion, Future Works, and Limitation}
In this study, we address the counter-intuition observed in MTSF where univariate models often outperform multivariate ones, despite the latter's ability to capture inter-series relationships. Our approach, \underline{C}onstructing \underline{A}uxiliary \underline{T}ime \underline{S}eries (CATS), enhances forecasting by generating ATS from OTS to represent inter-series relationships. CATS's efficacy is anchored in three key ATS principles — continuity, sparsity, and variability — along with the maintenance of the OTS shortcut. Our experiments reveal that CATS, even when paired with very simple predictor structures, significantly outperforms existing models in both strong and weak multivariate relationship scenarios. Furthermore, CATS offers a significant reduction in computational complexity and parameter size compared to other multivariate models, marking it as an easy-to-implement solution for MTSF.

For future works, CATS has the potential to be expanded into a general sequential modeling solution. With a wide range of ATS constructing methods and an increase in parameter size (such as using scalable Mixture of Experts for ATS construction, prediction, and projection), ATS can integrate very complex sequential relationships and perform efficient 2D attention-like operations. This gives it the potential to be transferred to fields like NLP and audio. CATS also has certain limitations: when constructing a fixed number of ATS, although it can address weak multivariate relationships through sparsity, if such relationships are too complex and require more ATS than the set number, the current CATS cannot expand automatically during training.


\section*{Impact Statement}

This paper advances Machine Learning by improving the accuracy and efficiency of multivariate time series forecasting with the proposed model, CATS. The method has beneficial applications like enhanced decision-making in domains such as finance and healthcare. While the societal impacts of this work are primarily positive, careful consideration and management of these technologies in sensitive applications are advised to prevent potential negative consequences.

\bibliography{example_paper}

@inproceedings{
lu2023arm,
title={{ARM}: Refining Multivariate Forecasting with Adaptive Temporal-Contextual Learning},
author={Jiecheng Lu and Xu Han and Shihao Yang},
booktitle={The Twelfth International Conference on Learning Representations},
year={2024},
url={https://openreview.net/forum?id=JWpwDdVbaM}
}

@inproceedings{nie2023time,
  title={A Time Series is Worth 64 Words: Long-term Forecasting with Transformers},
  author={Nie, Yuqi and Nguyen, Nam H and Sinthong, Phanwadee and Kalagnanam, Jayant},
  booktitle={The Eleventh International Conference on Learning Representations},
  year={2022}
}

@inproceedings{zeng2022transformers,
  title={Are transformers effective for time series forecasting?},
  author={Zeng, Ailing and Chen, Muxi and Zhang, Lei and Xu, Qiang},
  booktitle={Proceedings of the AAAI conference on artificial intelligence},
  volume={37},
  pages={11121--11128},
  year={2023}
}

@inproceedings{zhou2022fedformer,
    author = {Tian Zhou and
Ziqing Ma and
Qingsong Wen and
Xue Wang and
Liang Sun and
Rong Jin},
    bibsource = {dblp computer science bibliography, https://dblp.org},
    booktitle = {International Conference on Machine Learning, {ICML} 2022, 17-23 July
2022, Baltimore, Maryland, {USA}},
    editor = {Kamalika Chaudhuri and
Stefanie Jegelka and
Le Song and
Csaba Szepesv{\'{a}}ri and
Gang Niu and
Sivan Sabato},
    pages = {27268--27286},
    publisher = {{PMLR}},
    series = {Proceedings of Machine Learning Research},
    title = {FEDformer: Frequency Enhanced Decomposed Transformer for Long-term
Series Forecasting},
    url = {https://proceedings.mlr.press/v162/zhou22g.html},
    volume = {162},
    year = {2022}
}

@inproceedings{wu2022autoformer,
    author = {Haixu Wu and
Jiehui Xu and
Jianmin Wang and
Mingsheng Long},
    bibsource = {dblp computer science bibliography, https://dblp.org},
    booktitle = {Advances in Neural Information Processing Systems 34: Annual Conference
on Neural Information Processing Systems 2021, NeurIPS 2021, December
6-14, 2021, virtual},
    editor = {Marc'Aurelio Ranzato and
Alina Beygelzimer and
Yann N. Dauphin and
Percy Liang and
Jennifer Wortman Vaughan},
    pages = {22419--22430},
    title = {Autoformer: Decomposition Transformers with Auto-Correlation for Long-Term
Series Forecasting},
    year = {2021}
}

@inproceedings{zhou2021informer,
    author = {Haoyi Zhou and
Shanghang Zhang and
Jieqi Peng and
Shuai Zhang and
Jianxin Li and
Hui Xiong and
Wancai Zhang},
    bibsource = {dblp computer science bibliography, https://dblp.org},
    booktitle = {Thirty-Fifth {AAAI} Conference on Artificial Intelligence, {AAAI}
2021, Thirty-Third Conference on Innovative Applications of Artificial
Intelligence, {IAAI} 2021, The Eleventh Symposium on Educational Advances
in Artificial Intelligence, {EAAI} 2021, Virtual Event, February 2-9,
2021},
    pages = {11106--11115},
    publisher = {{AAAI} Press},
    title = {Informer: Beyond Efficient Transformer for Long Sequence Time-Series
Forecasting},
    url = {https://ojs.aaai.org/index.php/AAAI/article/view/17325},
    year = {2021}
}

@inproceedings{li2019enhancing,
    author = {Shiyang Li and
Xiaoyong Jin and
Yao Xuan and
Xiyou Zhou and
Wenhu Chen and
Yu{-}Xiang Wang and
Xifeng Yan},
    bibsource = {dblp computer science bibliography, https://dblp.org},
    booktitle = {Advances in Neural Information Processing Systems 32: Annual Conference
on Neural Information Processing Systems 2019, NeurIPS 2019, December
8-14, 2019, Vancouver, BC, Canada},
    editor = {Hanna M. Wallach and
Hugo Larochelle and
Alina Beygelzimer and
Florence d'Alch{\'{e}}{-}Buc and
Emily B. Fox and
Roman Garnett},
    pages = {5244--5254},
    title = {Enhancing the Locality and Breaking the Memory Bottleneck of Transformer
on Time Series Forecasting},
    year = {2019}
}

@inproceedings{lai2018modeling,
    author = {Guokun Lai and
Wei{-}Cheng Chang and
Yiming Yang and
Hanxiao Liu},
    bibsource = {dblp computer science bibliography, https://dblp.org},
    booktitle = {The 41st International {ACM} {SIGIR} Conference on Research {\&} Development
in Information Retrieval, {SIGIR} 2018, Ann Arbor, MI, USA, July 08-12,
2018},
    doi = {10.1145/3209978.3210006},
    editor = {Kevyn Collins{-}Thompson and
Qiaozhu Mei and
Brian D. Davison and
Yiqun Liu and
Emine Yilmaz},
    pages = {95--104},
    publisher = {{ACM}},
    title = {Modeling Long- and Short-Term Temporal Patterns with Deep Neural Networks},
    url = {https://doi.org/10.1145/3209978.3210006},
    year = {2018}
}

@inproceedings{kim2021reversible,
    author = {Taesung Kim and
Jinhee Kim and
Yunwon Tae and
Cheonbok Park and
Jang{-}Ho Choi and
Jaegul Choo},
    bibsource = {dblp computer science bibliography, https://dblp.org},
    booktitle = {The Tenth International Conference on Learning Representations, {ICLR}
2022, Virtual Event, April 25-29, 2022},
    publisher = {OpenReview.net},
    title = {Reversible Instance Normalization for Accurate Time-Series Forecasting
against Distribution Shift},
    url = {https://openreview.net/forum?id=cGDAkQo1C0p},
    year = {2022}
}

@article{box1974some,
    author = {Box, George EP and Jenkins, Gwilym M and MacGregor, John F},
    journal = {Journal of the Royal Statistical Society: Series C (Applied Statistics)},
    number = {2},
    pages = {158--179},
    publisher = {Wiley Online Library},
    title = {Some recent advances in forecasting and control},
    volume = {23},
    year = {1974}
}

@article{holt2004forecasting,
    author = {Holt, Charles C},
    journal = {International journal of forecasting},
    number = {1},
    pages = {5--10},
    publisher = {Elsevier},
    title = {Forecasting seasonals and trends by exponentially weighted moving averages},
    volume = {20},
    year = {2004}
}

@article{hochreiter1997long,
    author = {Hochreiter, Sepp and Schmidhuber, J{\"u}rgen},
    journal = {Neural computation},
    number = {8},
    pages = {1735--1780},
    publisher = {MIT press},
    title = {Long short-term memory},
    volume = {9},
    year = {1997}
}

@inproceedings{rangapuram2018deep,
    author = {Syama Sundar Rangapuram and
Matthias W. Seeger and
Jan Gasthaus and
Lorenzo Stella and
Yuyang Wang and
Tim Januschowski},
    bibsource = {dblp computer science bibliography, https://dblp.org},
    booktitle = {Advances in Neural Information Processing Systems 31: Annual Conference
on Neural Information Processing Systems 2018, NeurIPS 2018, December
3-8, 2018, Montr{\'{e}}al, Canada},
    editor = {Samy Bengio and
Hanna M. Wallach and
Hugo Larochelle and
Kristen Grauman and
Nicol{\`{o}} Cesa{-}Bianchi and
Roman Garnett},
    pages = {7796--7805},
    title = {Deep State Space Models for Time Series Forecasting},
    year = {2018}
}

@article{salinas2020deepar,
    author = {Salinas, David and Flunkert, Valentin and Gasthaus, Jan and Januschowski, Tim},
    journal = {International Journal of Forecasting},
    number = {3},
    pages = {1181--1191},
    publisher = {Elsevier},
    title = {DeepAR: Probabilistic forecasting with autoregressive recurrent networks},
    volume = {36},
    year = {2020}
}

@article{wen2017multi,
    author = {Wen, Ruofeng and Torkkola, Kari and Narayanaswamy, Balakrishnan and Madeka, Dhruv},
    journal = {ArXiv preprint},
    title = {A multi-horizon quantile recurrent forecaster},
    url = {https://arxiv.org/abs/1711.11053},
    volume = {abs/1711.11053},
    year = {2017}
}

@article{borovykh2017conditional,
    author = {Borovykh, Anastasia and Bohte, Sander and Oosterlee, Cornelis W},
    journal = {ArXiv preprint},
    title = {Conditional time series forecasting with convolutional neural networks},
    url = {https://arxiv.org/abs/1703.04691},
    volume = {abs/1703.04691},
    year = {2017}
}

@inproceedings{vanwavenet,
    author = {van den Oord, A{\"a}ron and Dieleman, Sander and Zen, Heiga and Simonyan, Karen and Vinyals, Oriol and Graves, Alex and Kalchbrenner, Nal and Senior, Andrew and Kavukcuoglu, Koray},
    booktitle = {9th ISCA Speech Synthesis Workshop},
    pages = {125--125},
    title = {WaveNet: A Generative Model for Raw Audio},
    year={2016}
}

@inproceedings{qin2017dual,
    author = {Yao Qin and
Dongjin Song and
Haifeng Chen and
Wei Cheng and
Guofei Jiang and
Garrison W. Cottrell},
    bibsource = {dblp computer science bibliography, https://dblp.org},
    booktitle = {Proceedings of the Twenty-Sixth International Joint Conference on
Artificial Intelligence, {IJCAI} 2017, Melbourne, Australia, August
19-25, 2017},
    doi = {10.24963/ijcai.2017/366},
    editor = {Carles Sierra},
    pages = {2627--2633},
    publisher = {ijcai.org},
    title = {A Dual-Stage Attention-Based Recurrent Neural Network for Time Series
Prediction},
    url = {https://doi.org/10.24963/ijcai.2017/366},
    year = {2017}
}

@article{analytics1020014,
    author = {Ma, Simin and Sun, Yan and Yang, Shihao},
    doi = {10.3390/analytics1020014},
    issn = {2813-2203},
    journal = {Analytics},
    number = {2},
    pages = {210--227},
    title = {Using Internet Search Data to Forecast COVID-19 Trends: A Systematic Review},
    url = {https://www.mdpi.com/2813-2203/1/2/14},
    volume = {1},
    year = {2022}
}

@inproceedings{zhang2023crossformer,
    author = {Zhang, Yunhao and Yan, Junchi},
    booktitle = {The Eleventh International Conference on Learning Representations},
    title = {Crossformer: Transformer utilizing cross-dimension dependency for multivariate time series forecasting},
    year = {2023}
}

@article{VAROriginal,
 ISSN = {00129682, 14680262},
 URL = {http://www.jstor.org/stable/1912017},
 author = {Christopher A. Sims},
 journal = {Econometrica},
 number = {1},
 pages = {1--48},
 publisher = {[Wiley, Econometric Society]},
 title = {Macroeconomics and Reality},
 urldate = {2024-01-16},
 volume = {48},
 year = {1980}
}

@InProceedings{Hu_2018_CVPR,
author = {Hu, Jie and Shen, Li and Sun, Gang},
title = {Squeeze-and-Excitation Networks},
booktitle = {Proceedings of the IEEE Conference on Computer Vision and Pattern Recognition (CVPR)},
month = {June},
year = {2018}
}

@inproceedings{wang2017residual,
  title={Residual attention network for image classification},
  author={Wang, Fei and Jiang, Mengqing and Qian, Chen and Yang, Shuo and Li, Cheng and Zhang, Honggang and Wang, Xiaogang and Tang, Xiaoou},
  booktitle={Proceedings of the IEEE conference on computer vision and pattern recognition},
  pages={3156--3164},
  year={2017}
}

@inproceedings{
dosovitskiy2021an,
title={An Image is Worth 16x16 Words: Transformers for Image Recognition at Scale},
author={Alexey Dosovitskiy and Lucas Beyer and Alexander Kolesnikov and Dirk Weissenborn and Xiaohua Zhai and Thomas Unterthiner and Mostafa Dehghani and Matthias Minderer and Georg Heigold and Sylvain Gelly and Jakob Uszkoreit and Neil Houlsby},
booktitle={International Conference on Learning Representations},
year={2021},
url={https://openreview.net/forum?id=YicbFdNTTy}
}

@inproceedings{liu2021swin,
  title={Swin transformer: Hierarchical vision transformer using shifted windows},
  author={Liu, Ze and Lin, Yutong and Cao, Yue and Hu, Han and Wei, Yixuan and Zhang, Zheng and Lin, Stephen and Guo, Baining},
  booktitle={Proceedings of the IEEE/CVF international conference on computer vision},
  pages={10012--10022},
  year={2021}
}

@misc{liu2023itransformer,
      title={iTransformer: Inverted Transformers Are Effective for Time Series Forecasting}, 
      author={Yong Liu and Tengge Hu and Haoran Zhang and Haixu Wu and Shiyu Wang and Lintao Ma and Mingsheng Long},
      year={2023},
      eprint={2310.06625},
      archivePrefix={arXiv},
      primaryClass={cs.LG}
}

@article{kaastra1996designing,
  title={Designing a neural network for forecasting financial and economic time series},
  author={Kaastra, Iebeling and Boyd, Milton},
  journal={Neurocomputing},
  volume={10},
  number={3},
  pages={215--236},
  year={1996},
  publisher={Elsevier}
}

@article{nihan1980use,
  title={Use of the Box and Jenkins time series technique in traffic forecasting},
  author={Nihan, Nancy L and Holmesland, Kjell O},
  journal={Transportation},
  volume={9},
  number={2},
  pages={125--143},
  year={1980},
  publisher={Springer}
}

@article{wen2022transformers,
  title={Transformers in time series: A survey},
  author={Wen, Qingsong and Zhou, Tian and Zhang, Chaoli and Chen, Weiqi and Ma, Ziqing and Yan, Junchi and Sun, Liang},
  journal={arXiv preprint arXiv:2202.07125},
  year={2022}
}

@InProceedings{KingBa15,
  author    = {Kingma, Diederik and Ba, Jimmy},
  booktitle = {International Conference on Learning Representations (ICLR)},
  title     = {Adam: A Method for Stochastic Optimization},
  year      = {2015},
  address   = {San Diega, CA, USA},
  optmonth  = {12},
}

@article{sun2023manifold,
  title={Manifold-constrained Gaussian process inference for time-varying parameters in dynamic systems},
  author={Sun, Yan and Yang, Shihao},
  journal={Statistics and Computing},
  volume={33},
  number={6},
  pages={142},
  year={2023},
  publisher={Springer}
}
\bibliographystyle{icml2024}

\newpage
\appendix
\onecolumn

\section{Datasets}\label{datasource}
We conduct our main MTSF experiments on 9 commonly used time series forecasting datasets, summarized as follows:

\paragraph{ETT Dataset\footnote{\url{https://github.com/zhouhaoyi/ETDataset}} \citep{zhou2021informer},} encompassing load and oil temperature data of electricity transformers, is recorded at 15-minute intervals from July 2016 to July 2018. It consists of four subsets: ETTm1, ETTm2, ETTh1, and ETTh2, representing two transformers (identified as 1 and 2) and two time resolutions (15 minutes and 1 hour). Each subset contains seven features related to oil and load of electricity transformers.

\paragraph{Electricity Dataset\footnote{\url{https://archive.ics.uci.edu/ml/datasets/ElectricityLoadDiagrams20112014}},} covering hourly electricity usage data of 321 consumers from 2012 to 2014, is frequently utilized in energy consumption forecasting and analysis.

\paragraph{Exchange Dataset\footnote{\url{https://github.com/laiguokun/multivariate-time-series-data}} \citep{lai2018modeling},} 
comprising daily exchange rates of eight countries from 1990 to 2016, offers insights into the global financial market dynamics.

\paragraph{Traffic Dataset\footnote{\url{http://pems.dot.ca.gov/}},}
 sourced from freeway sensors in the San Francisco Bay area, provides hourly road occupancy data from 2015 to 2016, serving as a key resource for traffic flow studies.

\paragraph{Weather Dataset\footnote{\url{https://www.bgc-jena.mpg.de/wetter/}}}
captures 21 weather variables, like temperature and humidity, recorded every 10 minutes throughout 2020, aiding in detailed meteorological studies.

\paragraph{Multi Dataset \citep{lu2023arm}} is generated based on a master random walk series. The first series is the master series, and the second to fifth series are derived by shifting the first series backward by 96, 192, 336, and 720 steps, respectively. The last three series are combinations of the first five series.

\paragraph{MultiX Dataset} is also generated based on a master random walk series. It comprises X series, where the first series is the master series, and each subsequent series is derived by shifting the previous series backward by $\frac{720}{X}$ steps. Before each shift, cumulative random noise is added to create a generative relationship with rich dependencies.

\subsubsection{Why we need MultiX 
dataset?}\label{multixdiscussion} Recent MTSF research has seen univariate models like PatchTST and DLinear achieve SOTA results in MTSF datasets, suggesting (a) potential design flaws in previous multivariate models and (b) possibly a lack of strong multivariate relationships in the current MTSF benchmark datasets. These univariate models inherently ignore inter-series/contextual information during inference, making them naturally more suited for datasets without inter-series relationships. However, their success is not sufficient to reject the existence or significance of contextual relationships in MTSF.

We believe CATS's performance improvement over other models stems from its effective modeling of inter-series relationships. We used the Multi/MultiX datasets, which emphasize inter-series relationships, to demonstrate CATS's effectiveness. However, since many datasets (like the ETTs) indeed lack significant inter-series dependencies, performance improvements from effectively modeling contextual information may appear marginal. We aim to highlight this through the CATS paper to prevent future MTSF research from misunderstanding due to benchmark dataset selection bias.

Among the commonly used benchmark datasets, ETTh1, ETTh2, ETTm1, ETTm2, Weather, and Exchange exhibit weak inter-series relationships. The Electricity dataset stands out as the only one that has more apparent inter-series dependencies. If a model enhances multivariate modeling capacities but only shows significant improvements on the Electricity dataset due to the lack of inter-series dependencies in others, it could be misleadingly deemed incremental upon evaluation. This seems unfair and misaligned with our actual expectations for MTSF research.

\section{Additional Explanation of Model Structure}\label{additionalexplanation}
\subsection{Continuity}\label{additionalcont}

\paragraph{Latent Separation of Inter-series Trend Effects} As aforementioned, in most real-world datasets, the univariate contribution tends to dominate in forecasting. The intra-series relationship here, in other words, the autocorrelation function (ACF) of the time series, governs their shape details. Considering a local temporal signal \( X(t) \), its ACF  
\( R_X(\tau) = \mathbb{E}[(X(t) - \mu_{X})(X(t + \tau) - \mu_{X})] \) 
describes the correlation of the signal with itself at different lags \( \tau \), where \( \mu_{X} \) is the mean of \( X(t) \). Even if a series is entirely generated by several other series, we can still establish its own shape-defining ACF based on the ACF of these other series. Considering a set of multivariate signals \( \{X_{\cdot}(t)\} \), assuming \( X_i(t) \) is completely generated by other local signals \( \{X_j(t)\}_{j \neq i} \) which can be explained by local ACF \(R_{X_j(t)}\), it can be represented as: \( X_i(t) = g_i(\{R_{X_j(t)}\}_{j \neq i}) \), where \( g_i \) is a generating function for local process $i$. The autocorrelation function \( R_{X_i}(\tau) \) of \( R_{X_j(t)} \) is then expressed as: 
\[R_{X_i}(\tau) = \mathbb{E}[(g_i(\{R_{X_j(t)}\}_{j \neq i}) - \mu_{X_i})(g_i(\{R_{X_j(t + \tau)}\}_{j \neq i}) - \mu_{X_i})]. \]
 In this generative system, since that only term that may not come from the series itself is $\mu_X$, if the trend in $X_{\cdot}(t)$ is controlled, meaning \( \mu_{X_{\cdot}} \) is fixed, even if $X_{\cdot}(t)$ is entirely generated by other series, its autocorrelation function \( R_{X_{\cdot}(t)}(\tau) \) will remain determinable in this system, which becomes non-local ACF \(R_{X_{\cdot}}(\tau)\). This actually means that the more $\mu_X$ we explain by using inter-series information (the intra-series effects of $\mu_X$ can equally be considered as inter-series effects without affecting our derivation), the more stable this system will be. Therefore, when we can explain more of the internal trend changes of a series using information beyond the series, the resulting more stable autocorrelation function can both enhance the reliability of forecasting and greatly reduce the complexity required by the main predictor. Consequently, the OTS part of CATS essentially computes all $X(t) = \text{Predict}(R_{X_{\cdot}(t)}(\tau), X(t-i))$, where $i \in {1, \ldots, \tau}$, using all stable ACFs. These ACFs, essentially latent functions within a neural network and not actual closed-form ACFs, enable the estimation of long-term detrend autocorrelations for all series by controlling the $\mu_X$ term. This approach separates the $\mu_X$ term latently in the CATS structure, thereby fixing the autocorrelation for each series and enabling the model to predict accurately through latent stable ACFs. In this scenario, the ATS's ability to extract changes in trend becomes key to model performance.

\paragraph{Spectral Density} We can use continuity loss to effectively enhance the ability of ATS to capture trends. Consider the spectral density of a local temporal signal \( X(t) \), represented as \[ S_X(f) = \int_{-\infty}^{\infty} R_X(\tau) e^{-i2\pi f\tau} d\tau , \] where \( f \) is the frequency. Let's consider an ATS signal \( A(t) \) generated based on \( X(t) \). The continuity loss reduces the differences between adjacent points in ATS. Consequently, compared to \( X(t) \), it imposes a penalty on the high-frequency components. This can be viewed as applying a low-pass filter \( H(f) \) to ATS. Therefore, the autocorrelation function of ATS, \( R_A(\tau) = R_X(\tau) * h(\tau) \), is influenced by this filtering, where \( * \) denotes convolution and \( h(\tau) \) is the inverse Fourier transform of \( H(f) \). Here, the spectral density of \( A(t) \) is \[ S_A(f) = \int_{-\infty}^{\infty} R_A(\tau) e^{-i2\pi f\tau} d\tau = S_X(f) \cdot H(f) .\] The \( H(f) \) reduces the high-frequency components in \( S_A(f) \), resulting in a lower amplitude of \( A(t) \) at high frequencies compared to \( X(t) \). This means that \( A(t) \) contains fewer short-term fluctuations or noise. As a result, the constructed ATS is smoother, reducing sensitivity to noise and focusing more on long-term trends.

\paragraph{Quadratic Variation and Denoising}
The role of Continuity in the CATS model can be understood from the perspectives of Quadratic Variation and Denoising. This theoretical viewpoint reveals how continuity aids in removing white noise from the ATS system and smoothens ATS predictions.

For a random process driven by Brownian motion (\( B_t \)), the quadratic variation \( [B]_a^b \) over the interval \([a, b]\) is equal to the length of the interval \( b - a \). This indicates that the quadratic variation of Brownian motion is closely related to the volatility along its path. According to its fundamental theorem, the quadratic variation of any differentiable function over any interval is zero, reflecting the smooth nature of its trajectory. Consider an ATS \( A(t) \), derived from an original time series \( X(t) \). If \( X(t) \) is driven by Brownian motion, then its quadratic variation over a certain interval is non-zero. However, by applying the Continuity principle, we aim to reduce the quadratic variation of \( A(t) \). Let \( A(t) \) be a continuous differentiable function, then its quadratic variation \( [A]_a^b \) over the interval \([a, b]\) is zero, which implies:

\[
[A]_a^b = \lim_{\|P\| \to 0} \sum_{i=1}^{n-1} (A(t_{i+1}) - A(t_i))^2 = 0
\]

where \( P \) is a partition of the interval \([a, b]\). In the CATS approach, by introducing the Continuity principle, we aim to decrease the volatility of the ATS, endowing it with properties akin to a differentiable function. This effectively removes short-term fluctuations, or white noise, from the time series, enabling the ATS to capture more of the long-term trends present in the OTS. 

\subsection{Time Series Predictors}\label{timeseriespredictor}
We introduce two predictor structures: the two-layer MLP (2L) and the independent linear layers (Ind) where weights are not shared between channels. The 2L predictor can be expressed as: \[ \Phi^{[\mathtt{2L}]}_{N+C}([A_I, X_I]) = \mathcal{W}_2 \cdot \mathtt{GELU}(\mathcal{W}_1 \cdot [A_I, X_I]^{\top} + \boldsymbol{\beta}_1)^{\top} + \boldsymbol{\beta}_2 \]
where \( \mathcal{W}_1, \mathcal{W}_2 \) are the weight matrices for the two layers, and \( \boldsymbol{\beta}_1, \boldsymbol{\beta}_2 \) are the corresponding biases. The IndLin predictor can be expressed as: \[
\Phi^{[\mathtt{IndLin}]}_{N+C}([A_I, X_I]^j) = \mathcal{W}^j \cdot [A_I, X_I]^{j \top} + \beta^{j}
\] where \( \mathcal{W}^j \) is the weight vector and \( \beta^j \) is the bias for the linear layer of channel $j$. The Mean predictor can be expressed as:
\[
\Phi^{[\mathtt{Mean}]}_{N+C}([A_I, X_I]^j) = \frac{1}{L_I}\sum_{t=1}^{L_I}[A_I, X_I]^{(t),j} \cdot \mathbf{1}_{L_P}
\]
where $\mathbf{1}_{L_P}$ is a ones vector with its length being $L_P$.

\section{Implementation Details}\label{implementationdetails}
\subsection{Hyper-parameter Settings}\label{hyperparametersettings}

We trained the CATS model using the Adam optimizer, employing a combined loss of Mean Squared Error (MSE) and a continuity loss term. The weigth of continuity loss $\beta_{cont}$ was set to 1. We used a learning rate of 0.00005, conducting training over 100 epochs with an early-stopping mechanism set at 30 steps. During these 100 epochs, a linear learning rate decay was applied. For all experiments, $L_I$ was set to 720. The training utilized a single Nvidia RTX 4090 GPU with a batch size of 32. Input processing strategies included random dropping and last-value demeaning as described in the main text.

A total of eight ATS constructors were consistently used, \textbf{\underline{detailed in Table \ref{additionalhyperparameter}}}. Datasets were classified into two categories: small datasets with $C < 16$ and large datasets with $C \geq 16$. The hyperparameter settings of ATS constructors for these two types of datasets are listed in Table \ref{additionalhyperparameter}. For the various 2-layer MLP structures in the text, the hidden layer dimension was set using a multiplier of $q$ times the input dimension, where the MLP ratio is $q$. For small datasets, $q=4$, and for large datasets, $q=8$. The 2-layer predictor (2L) included a dropout layer in the hidden layer, with a dropout rate of 0.75 for small datasets and 0.5 for large datasets. Independent linear and mean predictors did not require hyperparameter settings. For other MTSF baseline models used, we followed the settings from their original papers.

For model selection, following previous MTSF research practices, each dataset was divided into 70\% training, 10\% validation, and 20\% test sets. The best model was selected based on validation MSE loss, and MSE and MAE on the test set were calculated as reported results.

\begin{table}[hb]
\renewcommand{\arraystretch}{0.9}
\caption{Hyperparameter Settings of ATS Constructors}
\label{additionalhyperparameter}
\centering
\begin{threeparttable}
\begin{small}
\resizebox{\columnwidth}{!}{
\begin{tabular}{llll}
\toprule
 
Constructor &
Expression &
Settings ($C < 16$) &
Settings ($C \geq 16$)
\\
\midrule
Convolution (1) & $F^{[\mathtt{Conv}]}_1 (X_I) = \mathtt{Conv1d}_{[C \rightarrow n_1]}(X_I^{\top}, K, \lfloor \frac{K-1}{2} \rfloor, 1, 1)^{\top}$ & $K=49,n_1=8$ & $K=49,n_1=32$ \\
Convolution (2) & $F^{[\mathtt{Conv}]}_2 (X_I) = \mathtt{Conv1d}_{[C \rightarrow n_2]}(X_I^{\top}, K, \lfloor \frac{K-1}{2} \rfloor, 1, 1)^{\top}$ & $K=193,n_2=8$ & $K=193,n_1=32$ \\
Non-overlapping Convolution (1) & $F^{[\mathtt{NOConv}]}_3 (X_I) = \mathtt{Conv1d}_{[C \rightarrow n_3]}(X_I^{\top}, K, \lfloor \frac{E}{2} \rfloor, K, 1)^{\top}$ & $K=12,n_3=8$ & $K=12,n_3=32$ \\
Non-overlapping Convolution (2) & $F^{[\mathtt{NOConv}]}_4 (X_I) = \mathtt{Conv1d}_{[C \rightarrow n_4]}(X_I^{\top}, K, \lfloor \frac{E}{2} \rfloor, K, 1)^{\top}$ & $K=24,n_4=8$ & $K=24,n_4=32$ \\
Independent Convolution & $ F^{[\mathtt{IConv}]}_5 (X_I) = \mathtt{Conv1d}_{[C \rightarrow n_5]}(X_I^{\top}, K, \lfloor \frac{K-1}{2} \rfloor, 1, C)^{\top}$ & $K=49, n_5=1 \cdot C$ & $K=49, n_5=\max(1 \cdot C, \lceil \frac{32}{C} \rceil \cdot C)$ \\
Linear Projection & $F^{[\mathtt{Lin}]}_6 (X_I) = \mathtt{Conv1d}_{[C \rightarrow n_6]}(X_I^{\top}, 1, 0, 0, 1)^{\top}$ & $n_6=8$ & $n_6=32$ \\
Identity & $F^{[\mathtt{Id}]}_7 (X_I)=X_I$ & $n_7=C$ & $n_7=C$ \\
Embedding & $F^{[\mathtt{Emb}]}_8 (X_I)=\mathcal{W}^{[\mathtt{Emb}]}$ & $n_8=4$ & $n_8=16$ \\
\bottomrule

\end{tabular}
}
\end{small}
\end{threeparttable}
\vspace{-12pt}
\end{table}

\section{Additional Experiment Results}\label{addtionalresults}

Table \ref{fullresults} and Table \ref{fullmultiresults} present full comprehensive MTSF results under various forecasting horizons, highlighting the performance of CATS under both general real-world and strongly inter-related series datasets. Table \ref{fullstructureresults} shows the effectiveness of CATS with different time series predictors, while Table \ref{fullablationresult} focuses on the ablation studies of ATS principles and OTS shortcut. Additional performance comparisons of previous MTSF methods with and without CATS are detailed in Table \ref{addtionalMTSFresult}. Table \ref{ablationconstructor} and Table \ref{paramproportion} provide insights into the contributions of different ATS constructors and the parameter proportions for each module within CATS, respectively, demonstrating the model's robustness and efficiency.

\begin{figure}[hb]
	\centering
	\subfigure[Multi ATS ($L_P=96$)]{
		\begin{minipage}[b]{0.23\textwidth}
			\includegraphics[width=1\textwidth]{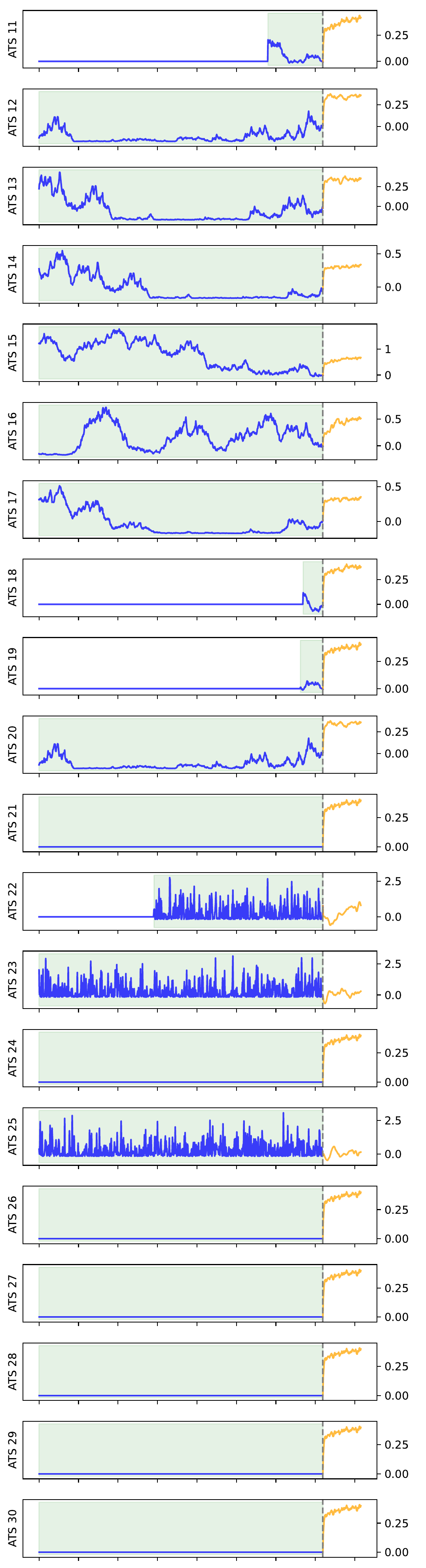}
		\end{minipage}
	}
    	\subfigure[Multi ATS ($L_P=192$)]{
    		\begin{minipage}[b]{0.23\textwidth}
   		 	\includegraphics[width=1\textwidth]{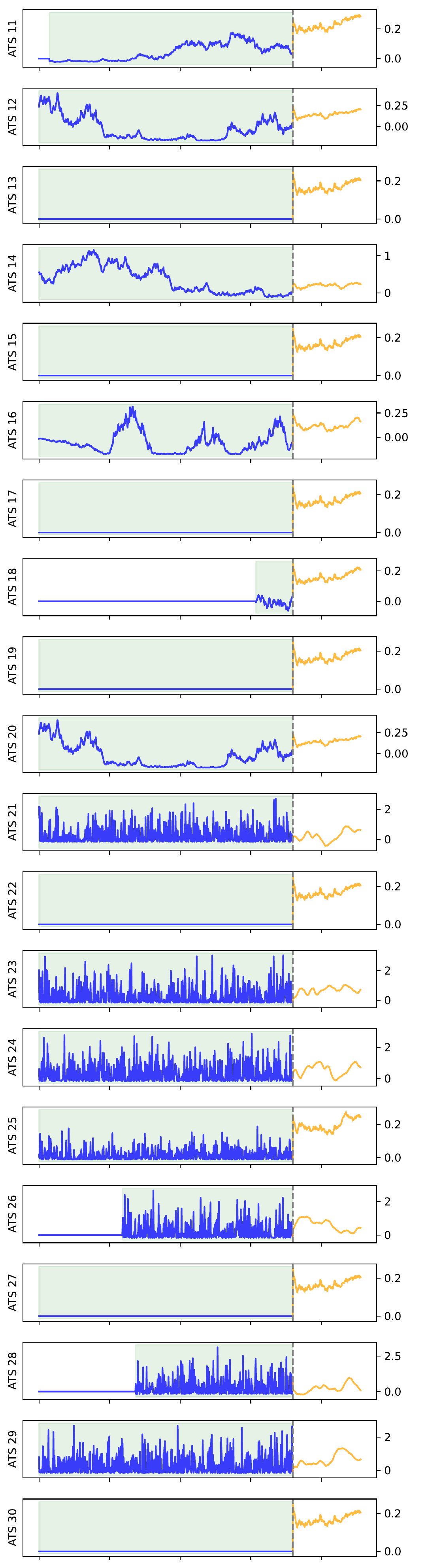}
    		\end{minipage}
    	}
        \subfigure[ETTh2 ATS ($L_P=336$)]{
    		\begin{minipage}[b]{0.23\textwidth}
   		 	\includegraphics[width=1\textwidth]{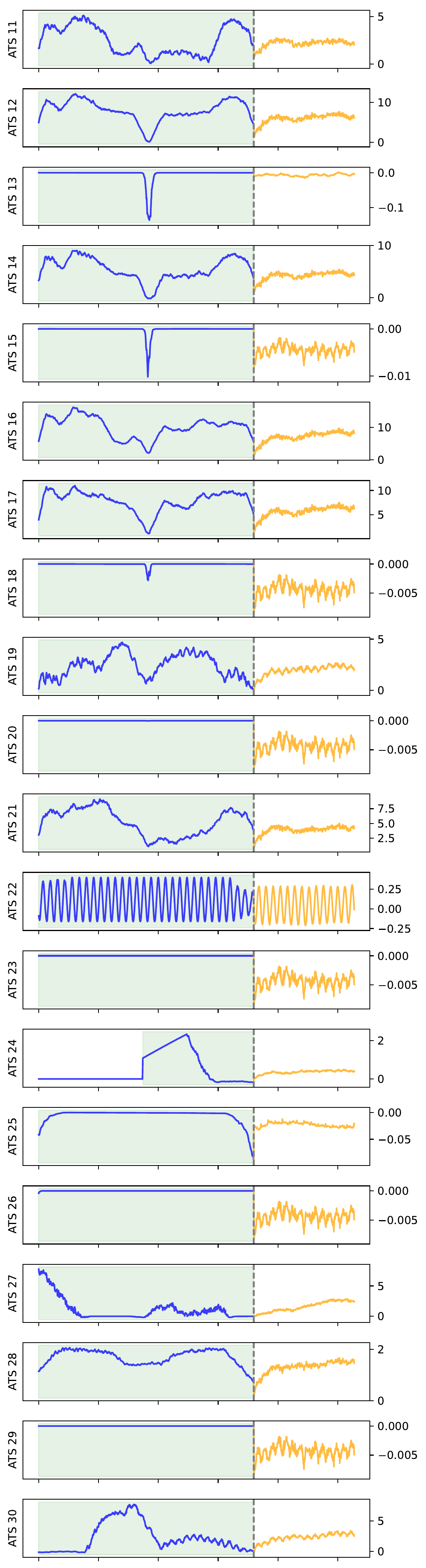}
    		\end{minipage}
    	}
        \subfigure[ETTh2 ATS ($L_P=720$)]{
    		\begin{minipage}[b]{0.23\textwidth}
   		 	\includegraphics[width=1\textwidth]{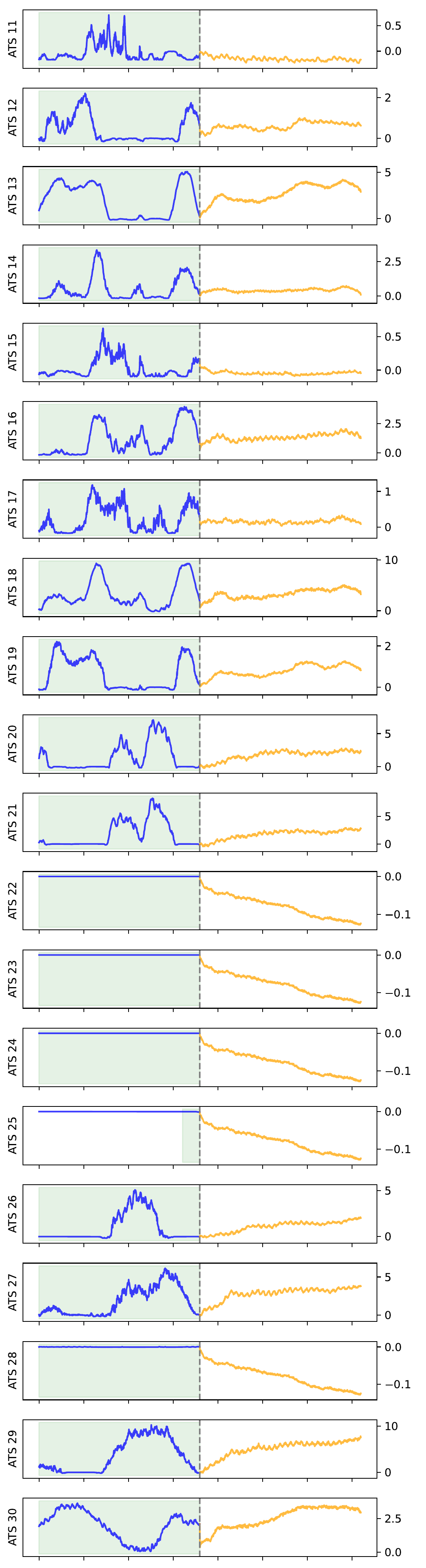}
    		\end{minipage}
    	}

     \subfigure[Multi OTS ($L_P=96$)]{
		\begin{minipage}[b]{0.23\textwidth}
			\includegraphics[width=1\textwidth]{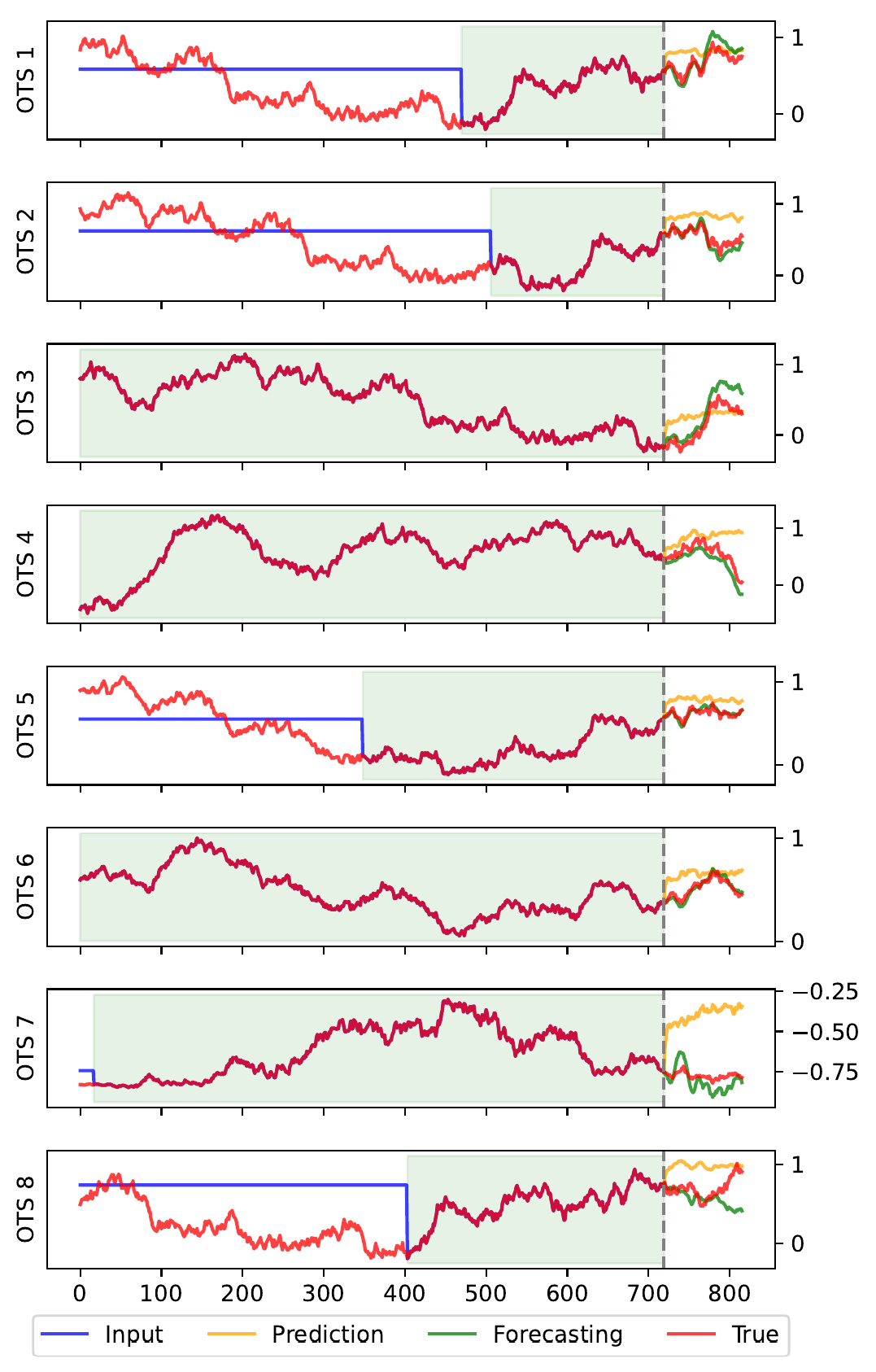}
		\end{minipage}
	}
    	\subfigure[Multi OTS ($L_P=192$)]{
    		\begin{minipage}[b]{0.23\textwidth}
   		 	\includegraphics[width=1\textwidth]{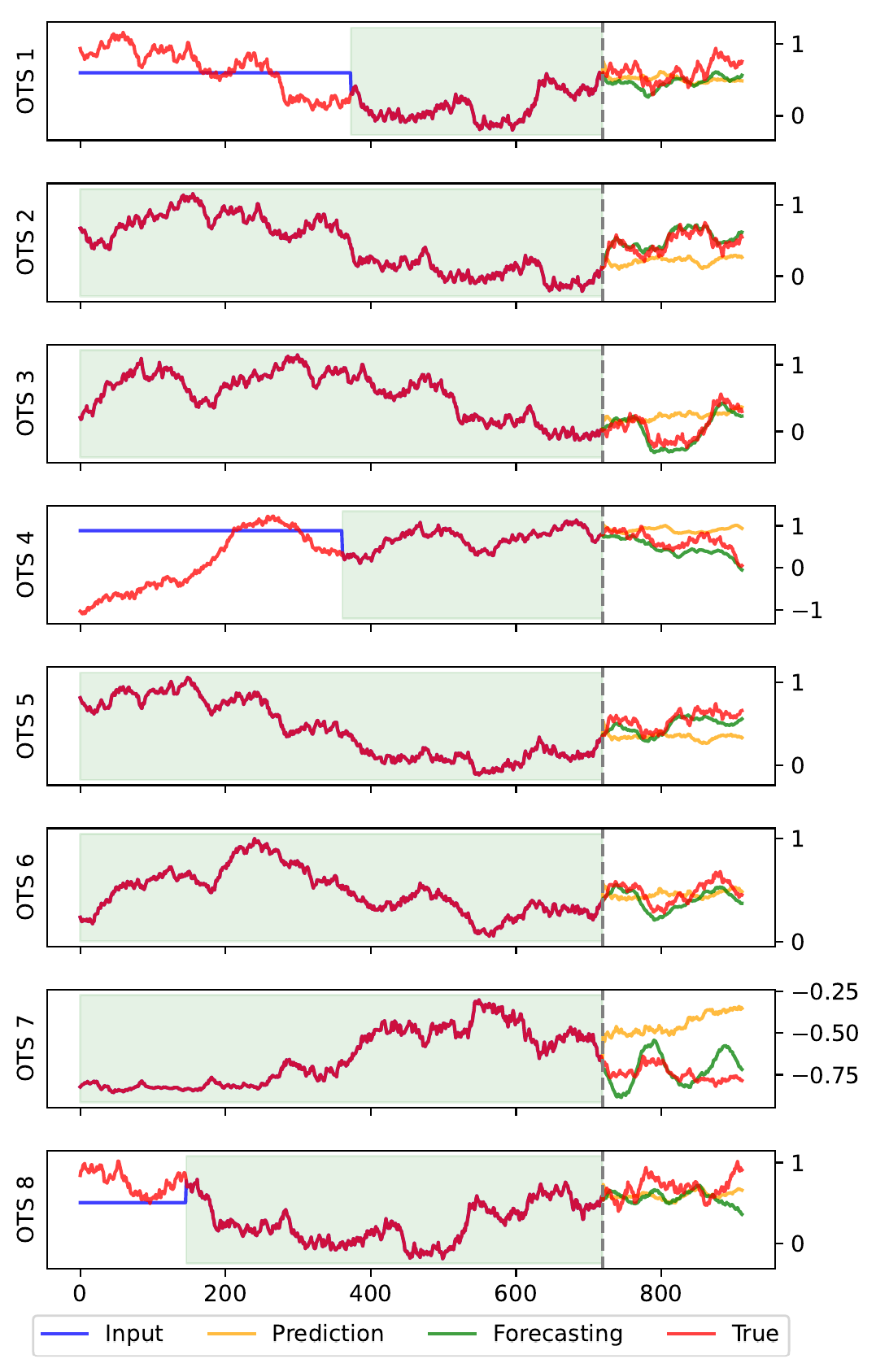}
    		\end{minipage}
    	}
        \subfigure[ETTh2 OTS ($L_P=336$)]{
    		\begin{minipage}[b]{0.23\textwidth}
   		 	\includegraphics[width=1\textwidth]{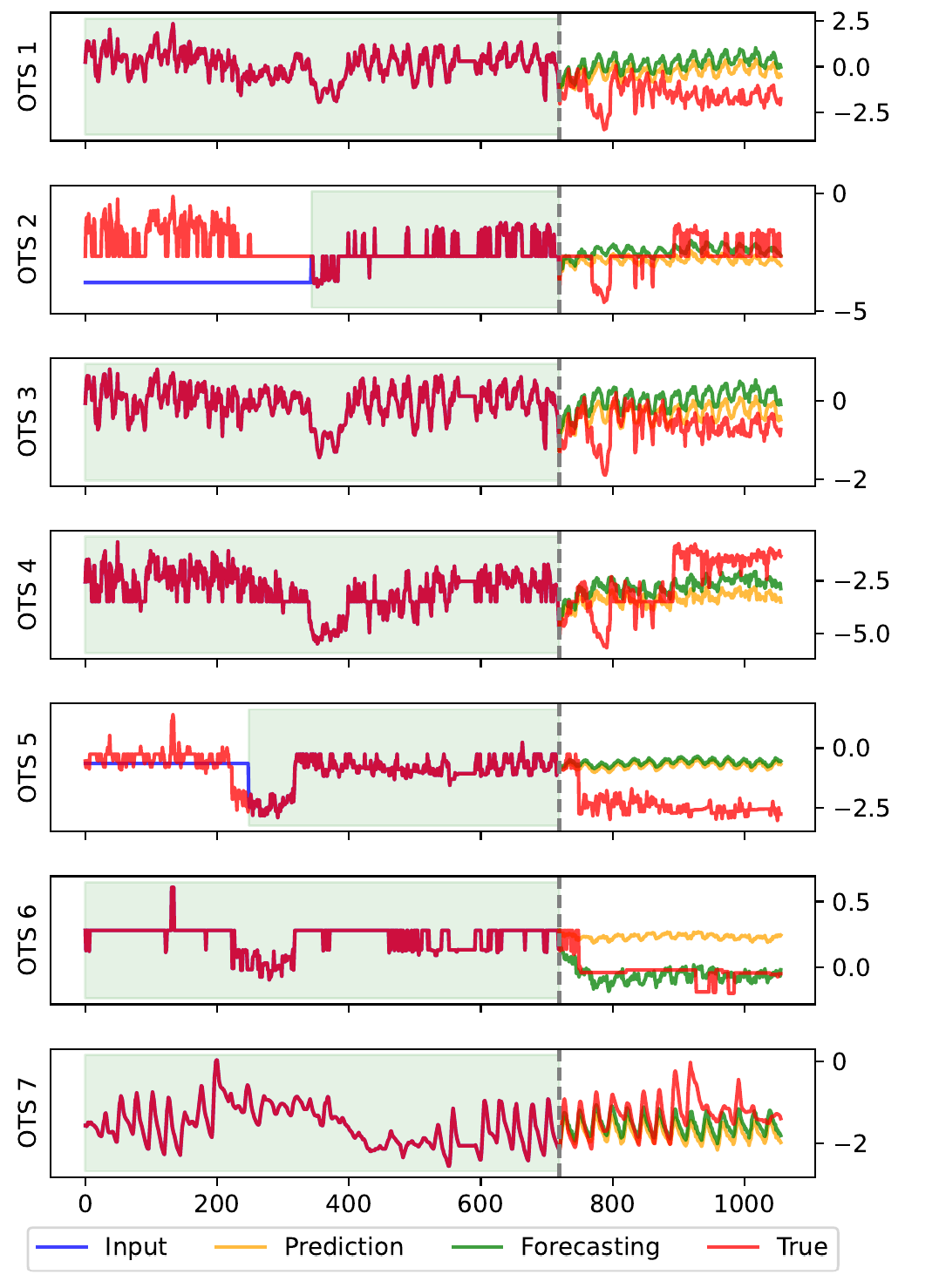}
    		\end{minipage}
    	}
        \subfigure[ETTh2 OTS ($L_P=720$)]{
    		\begin{minipage}[b]{0.23\textwidth}
   		 	\includegraphics[width=1\textwidth]{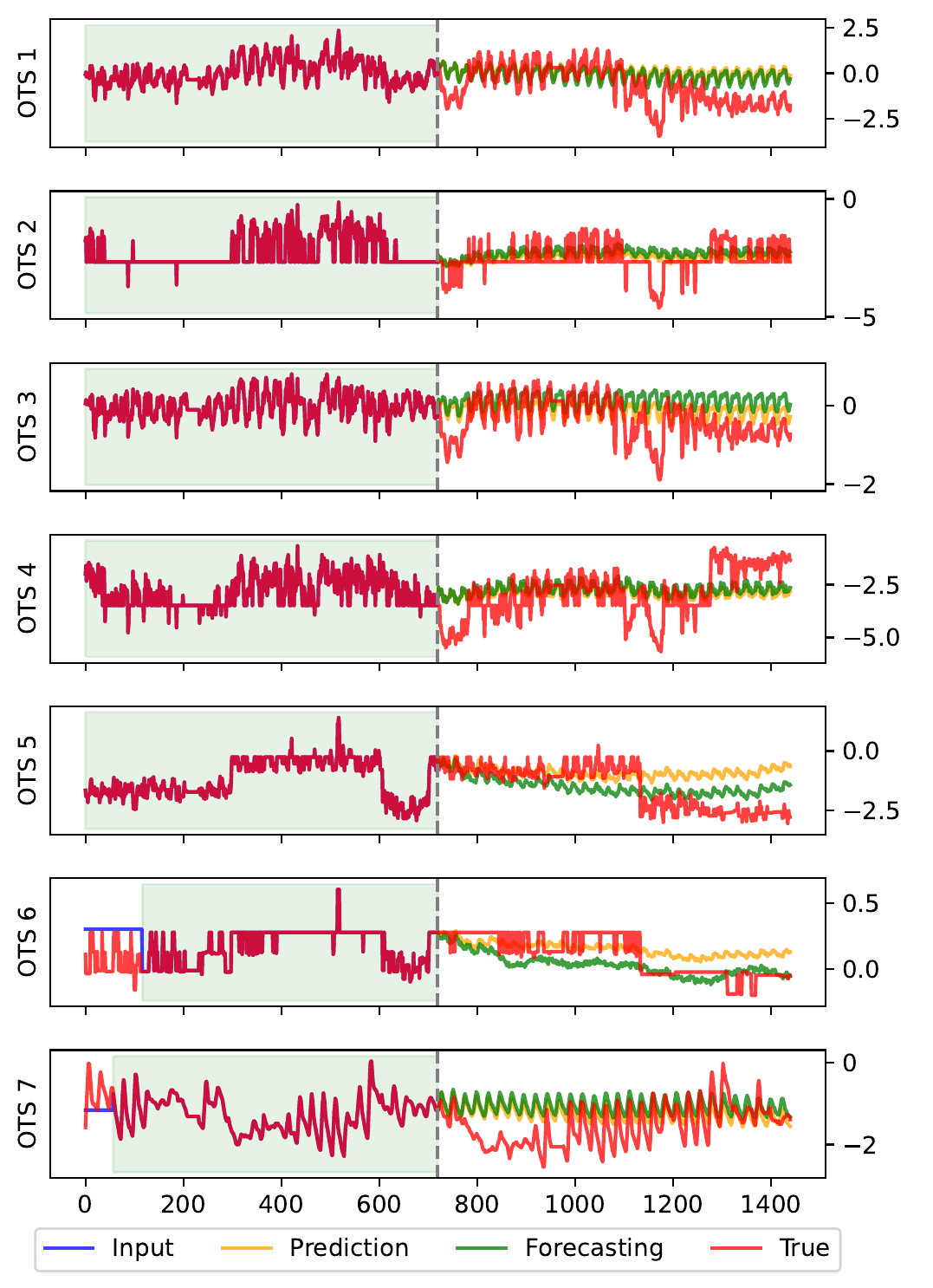}
    		\end{minipage}
    	}
     
	\caption{Visualization of CATS (2L) results for the Multi and ETTh2 datasets with one example datapoint for each plot. }
	\label{visualizationATS}
\end{figure}

Figure \ref{visualizationATS} provides the visualization of CATS (2L) results for the Multi and ETTh2 datasets. ATS from $A^{11}$ to $A^{30}$ and all the OTS are displayed. The red line represents the ground truth for each OTS channel. The blue line shows the predictor input for each ATS/OTS channel. \textbf{The yellow line indicates the first-stage ATS/OTS prediction. The green line is the final forecasting results, combining ATS and OTS predictions}. In the input domain, we use a green box to highlight the area remained after the temporal sparsity cutoff. Note that many input channels are turned into horizontal lines, which is caused by channel sparsity suppressing them; these series still have output prediction due to the bias in the 2L MLP.

\begin{table}[ht]
\renewcommand{\arraystretch}{0.5}
  \caption{Full MTSF results with forecasting horizons \( L_P \in \{96,192,336,720\} \). The best and second best results are in bold and underlined, respectively. Since CATS operates under the same experimental conditions as the models listed in the table, the results for baseline models are obtained from previous literature or from additional experiments conducted for datasets not reported in their literature. A last-value repeat strategy (naive model) is also included for comparison. For input length \( L_I \), CATS and ARM are fixed at 720, while other models select the best result from \( L_I \in \{96, 192, 336,720\} \) as their final result. }\label{fullresults}
  \centering
  \begin{threeparttable}
  \begin{small}
  \renewcommand{\multirowsetup}{\centering}
  \setlength{\tabcolsep}{8pt}
  \resizebox{\textwidth}{!}{   
  \begin{tabular}{lcccccccccccccccc}
    \toprule
    
    Models & \multicolumn{2}{c}{CATS (2L)} & \multicolumn{2}{c}{ARM} &  \multicolumn{2}{c}{PatchTST} & \multicolumn{2}{c}{DLinear}  & \multicolumn{2}{c}{FEDformer} & \multicolumn{2}{c}{Autoformer} & \multicolumn{2}{c}{Informer} & \multicolumn{2}{c}{Repeat}  \\
    
    \cmidrule(lr){2-3} \cmidrule(lr){4-5}\cmidrule(lr){6-7} \cmidrule(lr){8-9}\cmidrule(lr){10-11}\cmidrule(lr){12-13}\cmidrule(lr){14-15}\cmidrule(lr){16-17}
    
    Metric & MSE & MAE & MSE & MAE & MSE & MAE & MSE & MAE & MSE & MAE & MSE & MAE & MSE & MAE & MSE & MAE  \\

    \midrule

Electricity (96) & \textbf{0.125} & \underline{0.223} & \textbf{0.125} & \textbf{0.222} & \underline{0.129} & \textbf{0.222} & 0.140 & 0.237 & 0.193 & 0.308 & 0.201 & 0.317 & 0.274 & 0.368 & 1.588 & 0.946 \\
Electricity (192) & \textbf{0.142} & 0.241 & \textbf{0.142} & \textbf{0.239} & \underline{0.147} & \underline{0.240} & 0.153 & 0.249 & 0.201 & 0.315 & 0.222 & 0.334 & 0.296 & 0.386 & 1.595 & 0.950 \\
Electricity (336) & \underline{0.155} & \underline{0.253} & \textbf{0.154} & \textbf{0.251} & 0.163 & 0.259 & 0.169 & 0.267 & 0.214 & 0.329 & 0.231 & 0.338 & 0.300 & 0.394 & 1.617 & 0.961 \\
Electricity (720) & \textbf{0.174} & \textbf{0.273} & \underline{0.179} & \underline{0.275} & 0.197 & 0.290 & 0.203 & 0.301 & 0.246 & 0.355 & 0.254 & 0.361 & 0.373 & 0.439 & 1.647 & 0.975 \\
\midrule
ETTm1 (96) & \textbf{0.282} & \textbf{0.339} & \underline{0.287} & \underline{0.340} & 0.293 & 0.346 & 0.299 & 0.343 & 0.326 & 0.390 & 0.510 & 0.492 & 0.626 & 0.560 & 1.214 & 0.665 \\
ETTm1 (192) & \textbf{0.324} & \textbf{0.363} & \underline{0.328} & \underline{0.364} & 0.333 & 0.370 & 0.335 & 0.365 & 0.365 & 0.415 & 0.514 & 0.495 & 0.725 & 0.619 & 1.261 & 0.690 \\
ETTm1 (336) & \textbf{0.358} & \textbf{0.382} & \underline{0.364} & \underline{0.384} & 0.369 & 0.392 & 0.369 & 0.386 & 0.392 & 0.425 & 0.510 & 0.492 & 1.005 & 0.741 & 1.283 & 0.707 \\
ETTm1 (720) & \underline{0.414} & \underline{0.416} & \textbf{0.411} & \textbf{0.412} & 0.416 & 0.420 & 0.425 & 0.421 & 0.446 & 0.458 & 0.527 & 0.493 & 1.133 & 0.845 & 1.319 & 0.729 \\
\midrule
ETTm2 (96) & \textbf{0.158} & \textbf{0.248} & \underline{0.163} & \underline{0.254} & 0.166 & 0.256 & 0.167 & 0.260 & 0.203 & 0.287 & 0.255 & 0.339 & 0.365 & 0.453 & 0.266 & 0.328 \\
ETTm2 (192) & \textbf{0.211} & \textbf{0.285} & \underline{0.218} & \underline{0.290} & 0.223 & 0.296 & 0.224 & 0.303 & 0.269 & 0.328 & 0.281 & 0.340 & 0.533 & 0.563 & 0.340 & 0.371 \\
ETTm2 (336) & \textbf{0.264} & \textbf{0.322} & \underline{0.265} & \underline{0.324} & 0.274 & 0.329 & 0.281 & 0.342 & 0.325 & 0.366 & 0.339 & 0.372 & 1.363 & 0.887 & 0.412 & 0.410 \\
ETTm2 (720) & \textbf{0.340} & \textbf{0.371} & \underline{0.357} & \underline{0.382} & 0.362 & 0.385 & 0.397 & 0.421 & 0.421 & 0.415 & 0.422 & 0.419 & 3.379 & 1.388 & 0.521 & 0.465 \\
\midrule
ETTh1 (96) & \textbf{0.365} & \underline{0.396} & \underline{0.366} & \textbf{0.391} & 0.370 & 0.400 & 0.375 & 0.399 & 0.376 & 0.415 & 0.435 & 0.446 & 0.941 & 0.769 & 1.295 & 0.713 \\
ETTh1 (192) & \underline{0.404} & \underline{0.420} & \textbf{0.402} & 0.421 & 0.413 & 0.429 & 0.405 & \textbf{0.416} & 0.423 & 0.446 & 0.456 & 0.457 & 1.007 & 0.786 & 1.325 & 0.733 \\
ETTh1 (336) & 0.423 & \underline{0.437} & \textbf{0.421} & \textbf{0.431} & \underline{0.422} & 0.440 & 0.439 & 0.443 & 0.444 & 0.462 & 0.486 & 0.487 & 1.038 & 0.784 & 1.323 & 0.744 \\
ETTh1 (720) & \underline{0.441} & \underline{0.465} & \textbf{0.437} & \textbf{0.459} & 0.447 & 0.468 & 0.472 & 0.490 & 0.469 & 0.492 & 0.515 & 0.517 & 1.144 & 0.857 & 1.339 & 0.756 \\
\midrule
ETTh2 (96) & \textbf{0.259} & \textbf{0.327} & \underline{0.264} & \textbf{0.327} & 0.274 & \underline{0.337} & 0.289 & 0.353 & 0.332 & 0.374 & 0.332 & 0.368 & 1.549 & 0.952 & 0.432 & 0.422 \\
ETTh2 (192) & \textbf{0.309} & \textbf{0.368} & \underline{0.327} & \underline{0.374} & 0.341 & 0.382 & 0.383 & 0.418 & 0.407 & 0.446 & 0.426 & 0.434 & 3.792 & 1.542 & 0.534 & 0.473 \\
ETTh2 (336) & \textbf{0.329} & \textbf{0.383} & \underline{0.356} & 0.393 & \textbf{0.329} & \underline{0.384} & 0.448 & 0.465 & 0.400 & 0.447 & 0.477 & 0.479 & 4.215 & 1.642 & 0.591 & 0.508 \\
ETTh2 (720) & \textbf{0.358} & \underline{0.415} & \underline{0.371} & \textbf{0.408} & 0.379 & 0.422 & 0.605 & 0.551 & 0.412 & 0.469 & 0.453 & 0.490 & 3.656 & 1.619 & 0.588 & 0.517 \\
\midrule
Weather (96) & \textbf{0.138} & \textbf{0.191} & \underline{0.144} & \underline{0.193} & 0.149 & 0.198 & 0.176 & 0.237 & 0.217 & 0.296 & 0.266 & 0.336 & 0.300 & 0.384 & 0.259 & 0.254 \\
Weather (192) & \textbf{0.182} & \textbf{0.233} & \underline{0.189} & \underline{0.240} & 0.194 & 0.241 & 0.220 & 0.282 & 0.276 & 0.336 & 0.307 & 0.367 & 0.598 & 0.544 & 0.309 & 0.292 \\
Weather (336) & \textbf{0.230} & \textbf{0.274} & \underline{0.232} & \underline{0.280} & 0.245 & 0.282 & 0.265 & 0.319 & 0.339 & 0.380 & 0.359 & 0.395 & 0.578 & 0.523 & 0.377 & 0.338 \\
Weather (720) & \textbf{0.289} & \textbf{0.326} & \underline{0.296} & \underline{0.332} & 0.314 & 0.334 & 0.323 & 0.362 & 0.403 & 0.428 & 0.419 & 0.428 & 1.059 & 0.741 & 0.465 & 0.394 \\
\midrule
Traffic (96) & \textbf{0.347} & \textbf{0.243} & \underline{0.356} & \underline{0.247} & 0.360 & 0.249 & 0.410 & 0.282 & 0.587 & 0.366 & 0.613 & 0.388 & 0.719 & 0.391 & 2.723 & 1.079 \\
Traffic (192) & \textbf{0.368} & \textbf{0.256} & \underline{0.373} & \underline{0.258} & 0.379 & \textbf{0.256} & 0.423 & 0.287 & 0.604 & 0.373 & 0.616 & 0.382 & 0.696 & 0.379 & 2.756 & 1.087 \\
Traffic (336) & \textbf{0.379} & \underline{0.272} & \underline{0.383} & 0.274 & 0.392 & \textbf{0.264} & 0.436 & 0.296 & 0.621 & 0.383 & 0.622 & 0.337 & 0.777 & 0.420 & 2.791 & 1.095 \\
Traffic (720) & \textbf{0.421} & 0.297 & \underline{0.425} & \underline{0.294} & 0.432 & \textbf{0.286} & 0.466 & 0.315 & 0.626 & 0.382 & 0.660 & 0.408 & 0.864 & 0.472 & 2.811 & 1.097 \\
\midrule
Exchange (96) & \textbf{0.077} & \textbf{0.195} & \underline{0.078} & 0.197 & 0.087 & 0.207 & 0.081 & 0.203 & 0.148 & 0.278 & 0.197 & 0.323 & 0.847 & 0.752 & 0.081 & \underline{0.196} \\
Exchange (192) & \underline{0.157} & \underline{0.281} & \textbf{0.150} & \textbf{0.280} & 0.194 & 0.316 & \underline{0.157} & 0.293 & 0.271 & 0.380 & 0.300 & 0.369 & 1.204 & 0.895 & 0.167 & 0.289 \\
Exchange (336) & \underline{0.261} & \underline{0.374} & \textbf{0.252} & \textbf{0.367} & 0.351 & 0.432 & 0.305 & 0.414 & 0.460 & 0.500 & 0.509 & 0.524 & 1.672 & 1.036 & 0.305 & 0.396 \\
Exchange (720) & \underline{0.509} & \underline{0.542} & \textbf{0.486} & \textbf{0.535} & 0.867 & 0.697 & 0.643 & 0.601 & 1.195 & 0.841 & 1.447 & 0.941 & 2.478 & 1.310 & 0.823 & 0.681 \\
\midrule
Multi (96) & \textbf{0.022} & \textbf{0.092} & \underline{0.032} & \underline{0.125} & 0.072 & 0.202 & 0.067 & 0.190 & 0.117 & 0.261 & 0.162 & 0.313 & 0.092 & 0.219 & 0.068 & 0.189 \\
Multi (192) & \textbf{0.055} & \textbf{0.146} & \underline{0.063} & \underline{0.173} & 0.167 & 0.294 & 0.137 & 0.269 & 0.204 & 0.330 & 0.356 & 0.459 & 0.207 & 0.338 & 0.143 & 0.273 \\
Multi (336) & \textbf{0.155} & \textbf{0.282} & \underline{0.164} & \underline{0.286} & 0.314 & 0.405 & 0.238 & 0.355 & 0.313 & 0.402 & 0.572 & 0.705 & 0.284 & 0.414 & 0.264 & 0.369 \\
Multi (720) & \textbf{0.419} & \textbf{0.464} & \underline{0.450} & \underline{0.503} & 0.700 & 0.588 & 0.476 & 0.522 & 0.580 & 0.544 & 0.705 & 0.621 & 0.921 & 0.795 & 0.617 & 0.551 \\
 
    \bottomrule
  \end{tabular}  }
  
  \end{small}
  \end{threeparttable}
  \vspace{-12pt}
\end{table}

\begin{table}[ht]
  \caption{Extended MTSF results on datasets with strong inter-series relationship (\( L_P \in \{96,192,336,720\} \)). The construction method of MultiX datasets is detailed in section \ref{datasource}. For input length \( L_I \), CATS and ARM are fixed at 720, while other models select the best result from \( L_I \in \{336,720\} \) as their final result. The best and second best results are highlighted in bold and underlined, respectively.}\label{fullmultiresults}
  \centering
  \begin{threeparttable}
  \begin{small}
  \renewcommand{\multirowsetup}{\centering}
  \resizebox{\textwidth}{!}{   
  \begin{tabular}{lcccccccccccccccc}
    \toprule
    Models & \multicolumn{2}{c}{\textbf{CATS (2L)}} &  \multicolumn{2}{c}{ARM} &  \multicolumn{2}{c}{PatchTST} & \multicolumn{2}{c}{DLinear}  & \multicolumn{2}{c}{FEDformer} & \multicolumn{2}{c}{Autoformer} & \multicolumn{2}{c}{Informer} & \multicolumn{2}{c}{Repeat}  \\
    \cmidrule(lr){2-3} \cmidrule(lr){4-5}\cmidrule(lr){6-7} \cmidrule(lr){8-9}\cmidrule(lr){10-11}\cmidrule(lr){12-13}\cmidrule(lr){14-15}\cmidrule(lr){16-17}
    Metric & MSE & MAE & MSE & MAE & MSE & MAE & MSE & MAE & MSE & MAE & MSE & MAE & MSE & MAE & MSE & MAE  \\
    \midrule
Multi20 (96) & \textbf{0.003} & \textbf{0.042} & \underline{0.005} & \underline{0.051} & 0.014 & 0.091 & 0.013 & 0.088 & 0.033 & 0.130 & 0.038 & 0.151 & 0.015 & 0.089 & 0.014 & 0.091 \\
Multi20 (192) & \textbf{0.011} & \textbf{0.077} & \underline{0.014} & \underline{0.091} & 0.023 & 0.119 & 0.022 & 0.112 & 0.036 & 0.138 & 0.044 & 0.158 & 0.040 & 0.151 & 0.025 & 0.121 \\
Multi20 (336) & \textbf{0.015} & \textbf{0.091} & \underline{0.020} & \underline{0.110} & 0.027 & 0.129 & 0.027 & 0.124 & 0.053 & 0.171 & 0.055 & 0.168 & 0.086 & 0.201 & 0.032 & 0.138 \\
Multi20 (720) & \underline{0.038} & \textbf{0.122} & \textbf{0.035} & \underline{0.145} & 0.045 & 0.164 & \underline{0.038} & 0.149 & 0.109 & 0.258 & 0.093 & 0.237 & 0.198 & 0.354 & 0.044 & 0.165 \\
\midrule
Multi50 (96) & \textbf{0.004} & \textbf{0.045} & \underline{0.007} & \underline{0.061} & 0.014 & 0.092 & 0.013 & 0.089 & 0.031 & 0.128 & 0.037 & 0.151 & 0.017 & 0.094 & 0.014 & 0.091 \\
Multi50 (192) & \textbf{0.009} & \textbf{0.070} & \underline{0.014} & \underline{0.093} & 0.024 & 0.119 & 0.022 & 0.114 & 0.040 & 0.145 & 0.039 & 0.144 & 0.050 & 0.156 & 0.025 & 0.121 \\
Multi50 (336) & \textbf{0.018} & \textbf{0.097} & \underline{0.021} & \underline{0.113} & 0.028 & 0.131 & 0.027 & 0.123 & 0.048 & 0.161 & 0.046 & 0.158 & 0.144 & 0.325 & 0.032 & 0.139 \\
Multi50 (720) & 0.042 & 0.157 & \underline{0.040} & \underline{0.155} & 0.046 & 0.165 & \textbf{0.037} & \textbf{0.146} & 0.120 & 0.275 & 0.166 & 0.348 & 0.203 & 0.377 & 0.044 & 0.165 \\
\midrule
Multi100 (96) & \textbf{0.003} & \textbf{0.043} & \underline{0.006} & \underline{0.058} & 0.014 & 0.092 & 0.014 & 0.089 & 0.032 & 0.129 & 0.038 & 0.143 & 0.019 & 0.099 & 0.014 & 0.091 \\
Multi100 (192) & \textbf{0.008} & \textbf{0.066} & \underline{0.014} & \underline{0.091} & 0.024 & 0.120 & 0.022 & 0.114 & 0.038 & 0.142 & 0.037 & 0.139 & 0.046 & 0.153 & 0.025 & 0.121 \\
Multi100 (336) & \textbf{0.016} & \textbf{0.092} & \underline{0.021} & \underline{0.115} & 0.028 & 0.131 & 0.027 & 0.123 & 0.050 & 0.165 & 0.046 & 0.154 & 0.140 & 0.318 & 0.032 & 0.139 \\
Multi100 (720) & 0.045 & 0.168 & \underline{0.040} & \underline{0.154} & 0.046 & 0.165 & \textbf{0.038} & \textbf{0.148} & 0.114 & 0.267 & 0.356 & 0.536 & 0.243 & 0.424 & 0.045 & 0.165 \\
\midrule
Multi200 (96) & \textbf{0.003} & \textbf{0.039} & \underline{0.009} & \underline{0.075} & 0.013 & 0.088 & 0.014 & 0.091 & 0.034 & 0.133 & 0.030 & 0.133 & 0.014 & 0.088 & 0.015 & 0.092 \\
Multi200 (192) & \textbf{0.009} & \textbf{0.067} & \underline{0.017} & \underline{0.102} & 0.024 & 0.118 & 0.024 & 0.118 & 0.038 & 0.141 & 0.040 & 0.145 & 0.028 & 0.115 & 0.026 & 0.123 \\
Multi200 (336) & \textbf{0.010} & \textbf{0.102} & \underline{0.023} & \underline{0.118} & 0.027 & 0.126 & 0.028 & 0.127 & 0.055 & 0.175 & 0.084 & 0.225 & 0.043 & 0.155 & 0.033 & 0.141 \\
Multi200 (720) & 0.044 & 0.161 & \underline{0.043} & \underline{0.160} & 0.046 & 0.164 & \textbf{0.037} & \textbf{0.147} & 0.134 & 0.294 & 0.133 & 0.290 & 0.241 & 0.444 & 0.045 & 0.166 \\
    \bottomrule
  \end{tabular}  }
  
  \end{small}
  \end{threeparttable}
  \vspace{-12pt}
\end{table}

\begin{table}[ht]
  \caption{Results of utilizing CATS with different time series predictors. Input length $L_I$ are fixed as 720. The training strategies of last-value demeaning and random dropping are applied to all the baseline models. Better results are highlighted in bold. }\label{fullstructureresults}
  \centering
  \begin{threeparttable}
  \begin{small}
  \renewcommand{\multirowsetup}{\centering}
  \setlength{\tabcolsep}{5pt}
  \resizebox{\textwidth}{!}{   
  \begin{tabular}{l|cccc|cccc|cccc|cccc|cccc}
    \toprule
    \noalign{\vskip -0.5pt}
    Models & \multicolumn{2}{c}{2L} &  \multicolumn{2}{c|}{2L+CATS} & \multicolumn{2}{c}{DLinear} & \multicolumn{2}{c|}{DLinear+CATS} & \multicolumn{2}{c}{Autoformer} & \multicolumn{2}{c|}{Autoformer+CATS}  & \multicolumn{2}{c}{IndLin} & \multicolumn{2}{c|}{IndLin+CATS} & \multicolumn{2}{c}{Mean} & \multicolumn{2}{c}{Mean+CATS}   \\
    \noalign{\vskip -0.5pt}
    \cmidrule(lr){2-3} \cmidrule(lr){4-5}\cmidrule(lr){6-7} \cmidrule(lr){8-9}\cmidrule(lr){10-11}\cmidrule(lr){12-13}\cmidrule(lr){14-15}\cmidrule(lr){16-17}\cmidrule(lr){18-19}\cmidrule(lr){20-21}
    \noalign{\vskip -0.5pt}
    Metric & MSE & MAE & MSE & MAE & MSE & MAE & MSE & MAE & MSE & MAE & MSE & MAE & MSE & MAE & MSE & MAE & MSE & MAE & MSE & MAE  \\

    \midrule

Multi (96) & 0.082 & 0.209 & \textbf{0.022} & \textbf{0.092} & 0.070 & 0.193 & \textbf{0.024} & \textbf{0.073} & 0.345 & 0.427 & \textbf{0.034} & \textbf{0.119} & 0.066 & 0.187 & \textbf{0.015} & \textbf{0.064} & 0.439 & 0.488 & \textbf{0.054} & \textbf{0.175} \\
Multi (192) & 0.183 & 0.311 & \textbf{0.055} & \textbf{0.146} & 0.148 & 0.279 & \textbf{0.069} & \textbf{0.173} & 0.475 & 0.507 & \textbf{0.097} & \textbf{0.230} & 0.147 & 0.273 & \textbf{0.048} & \textbf{0.128} & 0.520 & 0.528 & \textbf{0.095} & \textbf{0.231} \\
Multi (336) & 0.357 & 0.426 & \textbf{0.155} & \textbf{0.282} & 0.276 & 0.379 & \textbf{0.176} & \textbf{0.292} & 0.613 & 0.585 & \textbf{0.253} & \textbf{0.384} & 0.283 & 0.376 & \textbf{0.140} & \textbf{0.232} & 0.647 & 0.581 & \textbf{0.188} & \textbf{0.322} \\
Multi (720) & 0.819 & 0.643 & \textbf{0.419} & \textbf{0.464} & 0.649 & 0.561 & \textbf{0.453} & \textbf{0.510} & 0.948 & 0.741 & \textbf{0.472} & \textbf{0.510} & 0.668 & 0.567 & \textbf{0.469} & \textbf{0.465} & 1.008 & 0.704 & \textbf{0.655} & \textbf{0.602} \\
\cmidrule(lr){1-1}\cmidrule(lr){2-5}\cmidrule(lr){6-9}\cmidrule(lr){10-13}\cmidrule(lr){14-17}\cmidrule(lr){18-21}
ETTh2 (96) & 0.275 & 0.336 & \textbf{0.259} & \textbf{0.327} & 0.270 & 0.341 & \textbf{0.267} & \textbf{0.338} & 0.363 & 0.410 & \textbf{0.304} & \textbf{0.373} & 0.279 & 0.342 & \textbf{0.273} & \textbf{0.339} & 0.388 & 0.425 & \textbf{0.322} & \textbf{0.386} \\
ETTh2 (192) & 0.345 & 0.381 & \textbf{0.309} & \textbf{0.368} & 0.331 & \textbf{0.375} & \textbf{0.326} & 0.379 & 0.381 & 0.423 & \textbf{0.345} & \textbf{0.398} & 0.335 & \textbf{0.378} & \textbf{0.329} & 0.379 & 0.403 & 0.434 & \textbf{0.372} & \textbf{0.416} \\
ETTh2 (336) & 0.372 & 0.405 & \textbf{0.329} & \textbf{0.383} & 0.370 & \textbf{0.401} & \textbf{0.363} & 0.408 & 0.384 & 0.434 & \textbf{0.364} & \textbf{0.411} & \textbf{0.365} & \textbf{0.405} & 0.367 & 0.406 & 0.406 & 0.438 & \textbf{0.391} & \textbf{0.430} \\
ETTh2 (720) & 0.398 & 0.433 & \textbf{0.358} & \textbf{0.415} & 0.394 & 0.434 & \textbf{0.383} & \textbf{0.432} & 0.433 & 0.474 & \textbf{0.389} & \textbf{0.447} & 0.381 & \textbf{0.423} & \textbf{0.374} & 0.427 & 0.434 & 0.461 & \textbf{0.426} & \textbf{0.456} \\
    \bottomrule
  \end{tabular}  }
  
  \end{small}
  \end{threeparttable}
  \vspace{-12pt}
\end{table}

\begin{table}[ht]
  \caption{Results of the ablation studies of the three ATS principles and OTS shortcut. Input length $L_I$ are fixed as 720. The training strategies of last-value demeaning and random dropping are applied to all the models. The best and second best results are highlighted in bold and underlined, respectively.}\label{fullablationresult}
  \centering
  \begin{threeparttable}
  \begin{small}
  \renewcommand{\multirowsetup}{\centering}
  \resizebox{\textwidth}{!}{   
  \begin{tabular}{lcccccccccccccccc}
    \toprule
    \multirow{2}{*}{Models} & \multicolumn{2}{c}{\multirow{2}{*}{CATS (2L)}} &  \multicolumn{2}{c}{\multirow{2}{*}{w/o Continuity}} &  \multicolumn{2}{c}{w/o Sparsity} & \multicolumn{2}{c}{w/o Sparsity}  & \multicolumn{2}{c}{w/o Sparsity} & \multicolumn{2}{c}{w/o Variability} & \multicolumn{2}{c}{w/o Variability} & \multicolumn{2}{c}{w/o OTS} \\
    & & & & & \multicolumn{2}{c}{(Channel)} & \multicolumn{2}{c}{(Temporal)} & \multicolumn{2}{c}{(Full)} & \multicolumn{2}{c}{(Pure Linear)} & \multicolumn{2}{c}{(Pure Conv)} & \multicolumn{2}{c}{Shortcut}
    \\
    \cmidrule(lr){2-3} \cmidrule(lr){4-5}\cmidrule(lr){6-7} \cmidrule(lr){8-9}\cmidrule(lr){10-11}\cmidrule(lr){12-13}\cmidrule(lr){14-15}\cmidrule(lr){16-17}
    Metric & MSE & MAE & MSE & MAE & MSE & MAE & MSE & MAE & MSE & MAE & MSE & MAE & MSE & MAE & MSE & MAE  \\
    \midrule
Multi (96) & \textbf{0.022} & \underline{0.092} & 0.031 & 0.122 & 0.026 & \textbf{0.091} & 0.028 & 0.107 & 0.035 & 0.107 & 0.043 & 0.151 & 0.035 & 0.126 & \underline{0.024} & 0.099 \\
Multi (192) & \textbf{0.055} & \textbf{0.146} & 0.081 & 0.197 & 0.089 & 0.191 & 0.074 & 0.178 & 0.082 & 0.195 & 0.096 & 0.217 & 0.089 & 0.210 & \underline{0.058} & \underline{0.153} \\
Multi (336) & \textbf{0.155} & \textbf{0.282} & \underline{0.178} & \underline{0.285} & 0.195 & 0.302 & 0.202 & 0.303 & 0.183 & 0.294 & 0.217 & 0.336 & 0.186 & 0.302 & 0.191 & 0.293 \\
Multi (720) & \textbf{0.419} & \textbf{0.464} & 0.634 & 0.566 & 0.710 & 0.570 & 0.519 & 0.573 & 0.731 & 0.600 & 0.595 & 0.563 & \underline{0.485} & \underline{0.497} & 0.562 & 0.523 \\
\midrule
ETTh2 (96) & \textbf{0.259} & \textbf{0.327} & 0.263 & \underline{0.329} & 0.267 & 0.334 & \underline{0.261} & \underline{0.329} & 0.270 & 0.338 & 0.265 & 0.330 & 0.266 & 0.334 & 0.288 & 0.353 \\
ETTh2 (192) & \textbf{0.309} & \textbf{0.368} & \underline{0.317} & 0.371 & 0.324 & 0.373 & 0.320 & 0.373 & 0.328 & 0.377 & \underline{0.317} & \underline{0.370} & 0.323 & 0.373 & 0.334 & 0.387 \\
ETTh2 (336) & \textbf{0.329} & \textbf{0.383} & \underline{0.338} & \underline{0.392} & 0.342 & 0.393 & 0.349 & 0.397 & 0.351 & 0.399 & 0.346 & 0.393 & 0.344 & 0.394 & 0.369 & 0.411 \\
ETTh2 (720) & \textbf{0.358} & \textbf{0.415} & 0.381 & 0.426 & \underline{0.362} & 0.451 & 0.376 & 0.423 & 0.372 & \underline{0.421} & 0.379 & 0.426 & 0.375 & 0.424 & 0.384 & 0.433 \\
    \bottomrule
  \end{tabular}  }
  
  \end{small}
  \end{threeparttable}
  \vspace{-12pt}
\end{table}

\begin{table}[ht]
  \caption{Additional results of other previous MTSF methods with and without CATS. Input length $L_I$ are fixed as 720. The best and second best results are highlighted in bold and underlined, respectively.}\label{addtionalMTSFresult}
  \centering
  \begin{threeparttable}
  \begin{small}
  \renewcommand{\multirowsetup}{\centering}
  \resizebox{\textwidth}{!}{   
  \begin{tabular}{lcccccccccccccccccccccc}
    \toprule
    \multirow{2}{*}{Models} & \multicolumn{2}{c}{2L} &  \multicolumn{2}{c}{\multirow{2}{*}{FITS}} &  \multicolumn{2}{c}{FITS} & \multicolumn{2}{c}{\multirow{2}{*}{NBeats}}  & \multicolumn{2}{c}{NBeats} & \multicolumn{2}{c}{\multirow{2}{*}{TiDE}} & \multicolumn{2}{c}{TiDE} & \multicolumn{2}{c}{\multirow{2}{*}{TimesNet}} & \multicolumn{2}{c}{TimesNet} & \multicolumn{2}{c}{\multirow{2}{*}{ARIMA}} & \multicolumn{2}{c}{\multirow{2}{*}{Repeat}}\\
    & \multicolumn{2}{c}{+CATS} & & & \multicolumn{2}{c}{+CATS} & & & \multicolumn{2}{c}{+CATS} & & & \multicolumn{2}{c}{+CATS} & & & \multicolumn{2}{c}{+CATS} & & & &
    \\
    \cmidrule(lr){2-3} \cmidrule(lr){4-5}\cmidrule(lr){6-7} \cmidrule(lr){8-9}\cmidrule(lr){10-11}\cmidrule(lr){12-13}\cmidrule(lr){14-15}\cmidrule(lr){16-17}\cmidrule(lr){18-19}\cmidrule(lr){20-21}\cmidrule(lr){22-23}
    Metric & MSE & MAE & MSE & MAE & MSE & MAE & MSE & MAE & MSE & MAE & MSE & MAE & MSE & MAE & MSE & MAE & MSE & MAE & MSE & MAE & MSE & MAE  \\
    \midrule

ETTh2 (96) & \textbf{0.259} & \textbf{0.327} & 0.271 & \underline{0.336} & \underline{0.269} & 0.340 & 0.643 & 0.560 & 0.300 & 0.363 & 0.273 & 0.339 & 0.270 & 0.338 & 0.388 & 0.422 & 0.315 & 0.381 & 0.392 & 0.407 & 0.432 & 0.422 \\
ETTh2 (192) & \textbf{0.309} & \textbf{0.368} & 0.330 & \underline{0.374} & \underline{0.322} & 0.381 & 1.226 & 0.738 & 0.387 & 0.420 & 0.334 & 0.378 & 0.329 & 0.388 & 0.407 & 0.430 & 0.394 & 0.432 & 0.503 & 0.467 & 0.534 & 0.473 \\
ETTh2 (336) & \textbf{0.329} & \textbf{0.383} & 0.353 & \underline{0.395} & \underline{0.349} & 0.401 & 1.347 & 0.787 & 0.424 & 0.448 & 0.356 & 0.400 & 0.351 & 0.405 & 0.399 & 0.438 & 0.391 & 0.435 & 0.572 & 0.512 & 0.591 & 0.508 \\
ETTh2 (720) & \underline{0.358} & \underline{0.415} & 0.378 & 0.422 & 0.375 & 0.430 & 1.411 & 0.839 & 0.436 & 0.469 & 0.385 & 0.431 & 0.380 & 0.431 & 0.425 & 0.456 & \textbf{0.350} & \textbf{0.407} & 0.622 & 0.541 & 0.588 & 0.517 \\
    \midrule
Multi (96) & \underline{0.022} & \underline{0.092} & 0.072 & 0.198 & 0.052 & 0.155 & 0.131 & 0.273 & \textbf{0.020} & \textbf{0.091} & 0.073 & 0.198 & 0.064 & 0.187 & 0.050 & 0.162 & 0.039 & 0.135 & 0.080 & 0.200 & 0.068 & 0.189 \\
Multi (192) & \textbf{0.055} & \textbf{0.146} & 0.153 & 0.288 & 0.141 & 0.242 & 0.304 & 0.425 & \underline{0.067} & \underline{0.152} & 0.145 & 0.285 & 0.114 & 0.250 & 0.093 & 0.219 & 0.087 & 0.211 & 0.190 & 0.298 & 0.143 & 0.273 \\
Multi (336) & \underline{0.155} & \underline{0.282} & 0.280 & 0.386 & 0.203 & 0.331 & 0.669 & 0.652 & \textbf{0.146} & \textbf{0.256} & 0.269 & 0.380 & 0.193 & 0.319 & 0.218 & 0.339 & 0.173 & 0.302 & 0.384 & 0.415 & 0.264 & 0.369 \\
Multi (720) & \textbf{0.419} & \textbf{0.464} & 0.684 & 0.587 & 0.513 & 0.531 & 1.368 & 1.001 & \underline{0.480} & \underline{0.493} & 0.632 & 0.559 & 0.547 & 0.551 & 0.627 & 0.569 & 0.542 & 0.539 & 1.064 & 0.653 & 0.617 & 0.551 \\

    \bottomrule
  \end{tabular}  }
  
  \end{small}
  \end{threeparttable}
  \vspace{-12pt}
\end{table}

\begin{table}[tb]
\caption{Additional ablation studies of ATS constructors. "Full" represents the default CATS configuration (2L), with multiple baseline models added for comparison after the Pure Conv ablation. Different ATS constructors may be suited to different datasets, and some datasets might not require certain ATS constructors (as shown in the table below for ETTh2's Conv, and ETTm2's Id and Emb). However, thanks to the sparsity module, there's no need for manual experimentation to select the most appropriate ATS constructors. The incomplete CATS models are still enough to beat previous SOTAs.}
\label{ablationconstructor}
\centering
\begin{threeparttable}
\begin{small}
\resizebox{\columnwidth}{!}{
\begin{tabular}{lcccccccccccccc}
\toprule
 Models & Full & w/o Conv & w/o NOConv & w/o Iconv & w/o Lin & w/o Id & w/o Emb & Pure Lin & Pure Conv & ARM & PatchTST & DLinear & Autoformer & Informer \\
\midrule
ETTh2 (196) & 0.310 & 0.310 & 0.317 & 0.314 & 0.315 & 0.316 & 0.315 & 0.317 & 0.323 & 0.327 & 0.341 & 0.383 & 0.426 & 3.792 \\
ETTm2 (196) & 0.218 & 0.219 & 0.219 & 0.221 & 0.220 & 0.218 & 0.218 & 0.220 & 0.224 & 0.218 & 0.223 & 0.224 & 0.281 & 0.533 \\
\bottomrule
\end{tabular}
}
\end{small}
\end{threeparttable}
\vspace{-12pt}
\end{table}

\begin{table}[tb]
\renewcommand{\arraystretch}{0.9}
\caption{The proportion of parameter sizes for each module within CATS for the ETTh2 setting ($C=7$, $L_I=720$, $L_P=96$) and the Electricity setting ($C=321$, $L_I=720$, $L_P=96$), calculated using PyTorch ".numel()" method.}
\label{paramproportion}
\centering
\begin{threeparttable}
\begin{small}
\resizebox{\columnwidth}{!}{
\begin{tabular}{lccccccccc}
\toprule
 Module & Predictor (2L) & $F^{[\mathtt{NOConv}]}$ & Temporal Sparsity & Channel Sparsity & $F^{[\mathtt{Conv}]}$	 & $F^{[\mathtt{Emb}]}$ & $F^{[\mathtt{IConv}]}$ & Output Projection & $F^{[\mathtt{Lin}]}$ \\
 \midrule
Params (ETTh2) & 2353056 & 63648 & 50469 & 50434 & 21312 & 2880 & 700 & 518 & 96 \\
Params\% (ETTh2) & 92.53\% & 2.50\% & 1.98\% & 1.98\% & 0.84\% & 0.11\% & 0.03\% & 0.02\% & 0.00\% \\
Params (Electricity) & 2353056 & 1872288 & 276863 & 2208556 & 629216 & 2880 & 700 & 2716 & 2608 \\
Params\%  (Electricity) & 32.02\% & 25.48\% & 3.77\% & 30.05\% & 8.56\% & 0.04\% & 0.01\% & 0.04\% & 0.04\% \\
\bottomrule
\end{tabular}
}
\end{small}
\end{threeparttable}
\vspace{-12pt}
\end{table}

\end{document}